\documentclass[runningheads]{llncs}
\usepackage{ifpdf}
\usepackage{ifxetex}
\usepackage{microtype}
\usepackage{comment}
\usepackage{animate}
\usepackage{tablefootnote}
\usepackage{amsmath,amssymb} 
\usepackage{color}
\usepackage{amsmath,amssymb} 
\usepackage{array}
\usepackage{color}
\usepackage{multirow}
\usepackage{makecell, booktabs, caption}
\usepackage[position=below]{subfig}
\usepackage[ruled,vlined]{algorithm2e}
\usepackage{pifont}
\usepackage{graphicx,subfig}
\usepackage[width=122mm,left=12mm,paperwidth=146mm,height=193mm,top=12mm,paperheight=217mm]{geometry}

\usepackage[colorlinks = true,
linkcolor = blue,
urlcolor  = blue,
citecolor = blue,
anchorcolor = blue]{hyperref}
\newcommand{\losscycle}{\mathcal{L}_{c}} 
\newcommand{\losstotal}{\mathcal{L}} 
\newcommand{\lossmatch}{\mathcal{L}_{m}} 
\newcommand{\losssim}{\mathcal{L}_{rec}} 
\newcommand{\matcheta}{\lambda} 
\newcommand{\flowtheta}{\mu} 
\newcolumntype{P}[1]{>{\centering\arraybackslash}p{#1}}

\newcommand{\Is}{I_s} 
\newcommand{\It}{I_t} 
\newcommand{\kernel}{K} 
\newcommand{\flow}{\mathbf{F}} 
\newcommand{\flowIm}{F} 
\newcommand{\match}{M} 
\newcommand{\sourceFeat}{f_s} 
\newcommand{\targetFeat}{f_t} 
\newcommand{\spatialW}{W} 
\newcommand{\spatialH}{H} 
\newcommand{\simi}{s} 

\begin{document}
	\pagestyle{headings}
	\mainmatter
	\def\ECCVSubNumber{2623}  
	
	\title{RANSAC-Flow: generic two-stage \\
		image alignment}

	%
	\author{Xi Shen \inst{1} \and
		Fran\c cois Darmon \inst{1,2} \and
		Alexei A. Efros \inst{3} \and
		Mathieu Aubry \inst{1} 
	}
	\authorrunning{X. Shen et al.}
	\institute{LIGM (UMR 8049) - Ecole des Ponts, UPE
		\and Thales Land and Air Systems
		\and
		UC Berkeley} 

	\maketitle
	
	\begin{abstract}
		This paper considers the generic problem of dense alignment between two images, whether they be two frames of a video, two widely different views of a scene, two paintings depicting similar content, etc. Whereas each such task is typically addressed with a domain-specific solution, we show that a simple unsupervised approach performs surprisingly well across a range of tasks. 
		Our main insight is that parametric and non-parametric alignment methods have complementary strengths.
		We propose a two-stage process: first, a feature-based parametric coarse alignment using one or more homographies, followed by non-parametric fine pixel-wise alignment. Coarse alignment is performed using RANSAC on off-the-shelf deep features.
		Fine alignment is learned in an unsupervised way by a deep network which optimizes a standard structural similarity metric (SSIM) between the two images, plus cycle-consistency. 
		Despite its simplicity, our method shows competitive results on a range of tasks and datasets, including unsupervised optical flow on {\sc KITTI}, dense correspondences on {\sc Hpatches}, two-view geometry estimation on {\sc YFCC100M}, localization on {\sc Aachen Day-Night}, and, for the first time, fine alignment of artworks on the {\sc Brughel dataset}. Our code and data are available at \url{http://imagine.enpc.fr/~shenx/RANSAC-Flow/}.
		
		
	\end{abstract}

	\section{Introduction}
	
	\captionsetup[subfloat]{labelformat=empty,farskip=1pt,captionskip=1pt}
	\begin{figure}[!t]
		\begin{center}	
			\includegraphics[width=1.0\textwidth]{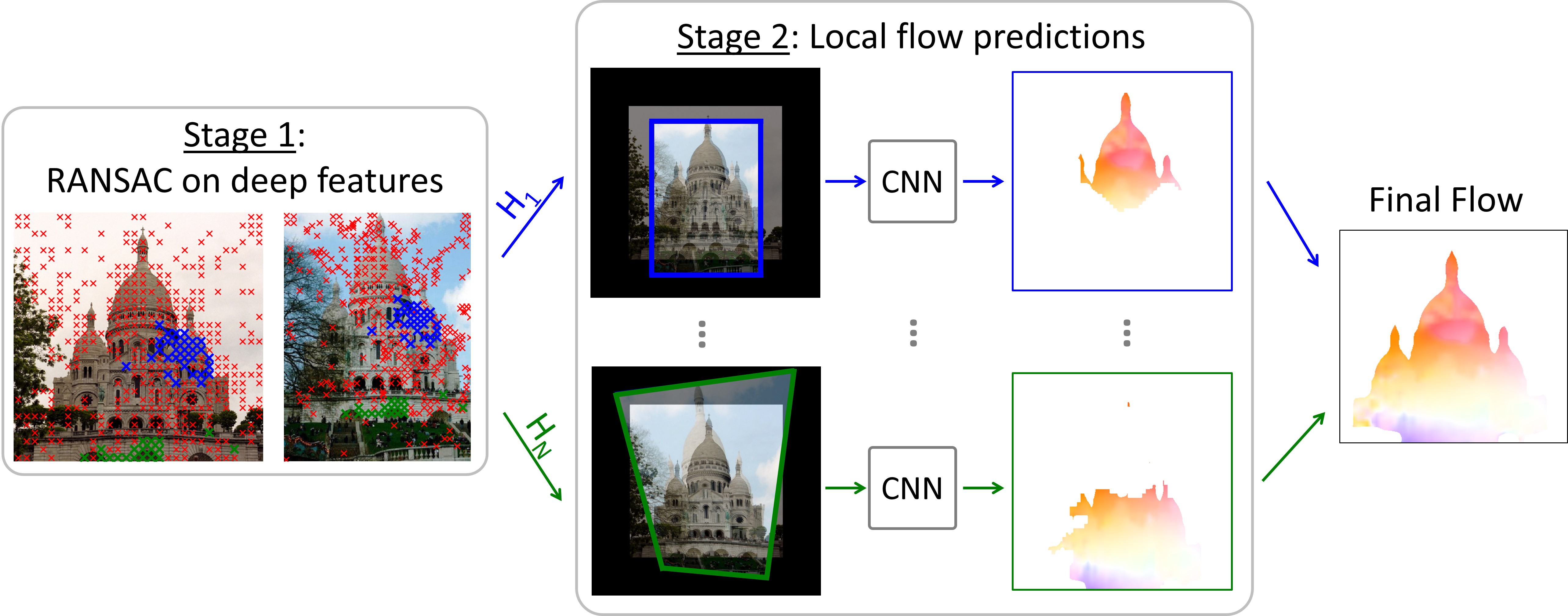}
		\end{center}
		\caption{{\bf Overview of RANSAC-Flow.} Stage 1: given a pair of images, we compute sparse correspondences (using off-the-shelf deep features), use RANSAC to estimate a homography, and warp second image using it. Stage 2: given two coarsely aligned images, our self-supervised fine flow network generates flow predictions in the matchable region.  To compute further homographies, we can remove matched correspondences, and iterate the process. }
		\label{fig:pipline}
	\end{figure}
	
	\begin{figure}[t]
		\begin{minipage}{0.49\textwidth}
			\begin{center}
				\subfloat{\includegraphics[width=0.7\textwidth, height = 0.2\textwidth]{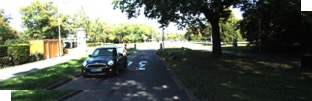}}\\
				\subfloat[(a) Optical flow estimation.]{\includegraphics[width=0.7\textwidth, height = 0.2\textwidth]{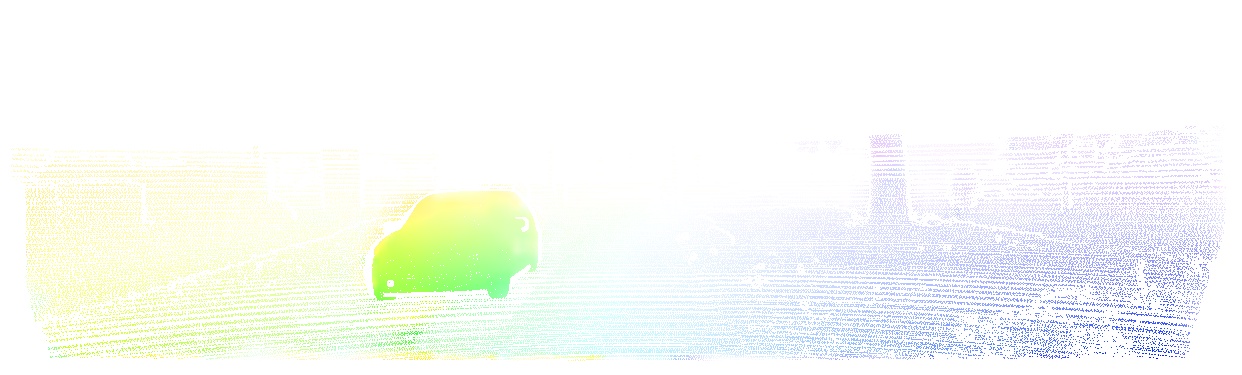}}
			\end{center}
		\end{minipage}
		\begin{minipage}{0.49\textwidth}
			\begin{center}
				\subfloat[(b) Visual localization.]{\includegraphics[width=0.95\textwidth, height = 0.4\textwidth]{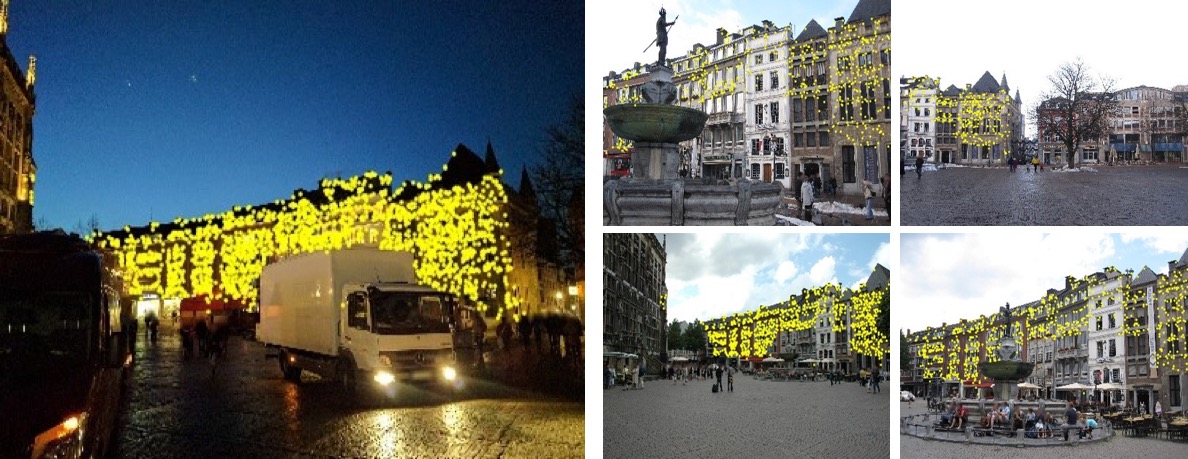}}
			\end{center}
		\end{minipage}
		\begin{minipage}{0.49\textwidth}
			\begin{center}
				\subfloat[(c) 2-view geometry estimation.]{\includegraphics[width=0.90\textwidth, height = 0.45\textwidth]{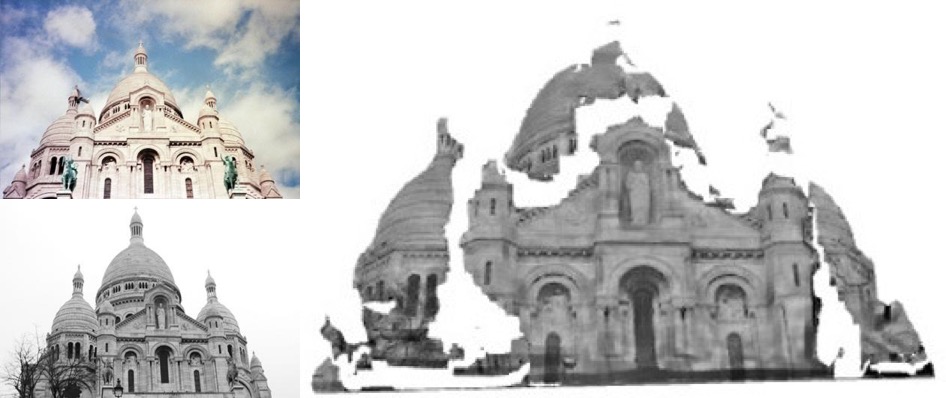}}
			\end{center}
		\end{minipage}
		\begin{minipage}{0.49\textwidth}
			\begin{center}
				\subfloat[(d) Dense image alignment.]{\includegraphics[width=0.90\textwidth, height = 0.45\textwidth]{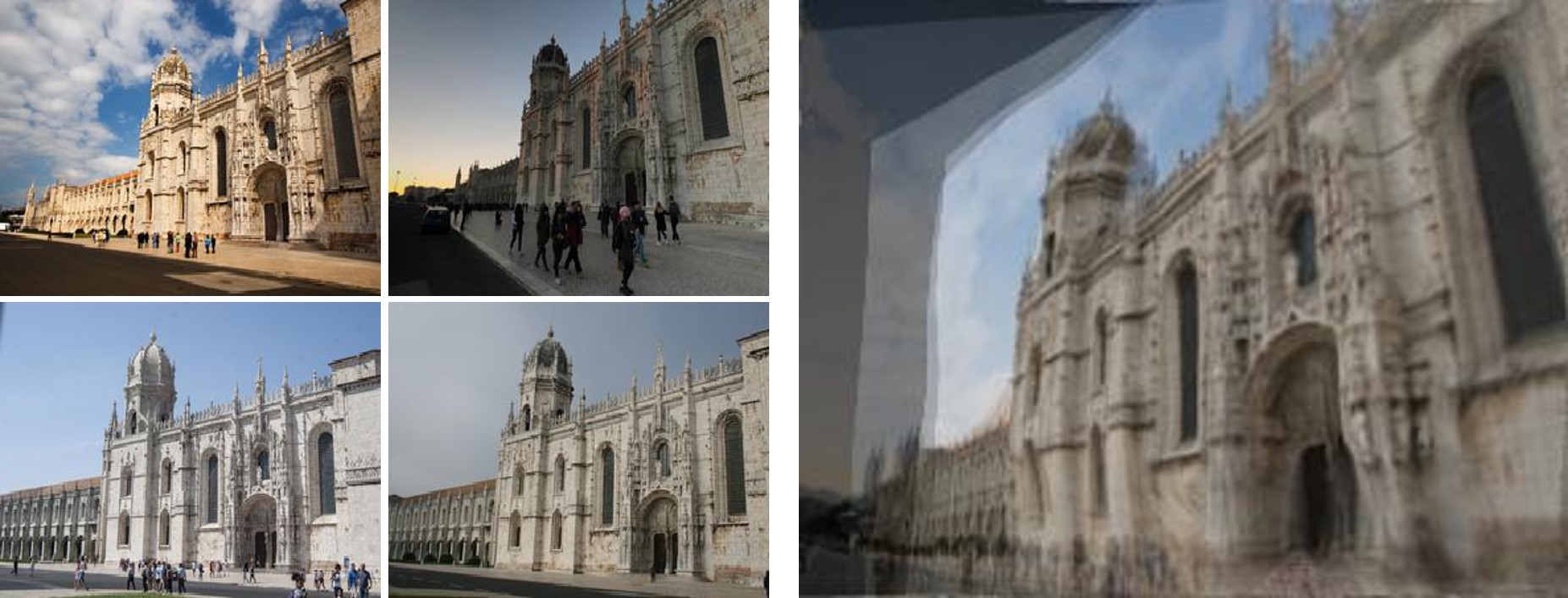}}
			\end{center}
		\end{minipage}
		\begin{minipage}{0.49\textwidth}
			\begin{center}
				\subfloat[(e) Artwork alignment.]{\includegraphics[width=0.97\textwidth,height=0.5\textwidth]{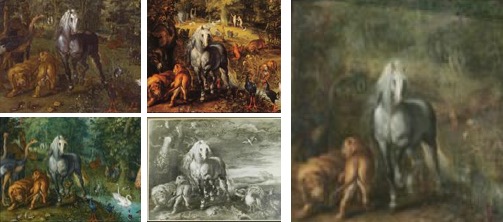}}
			\end{center}
		\end{minipage}
		\begin{minipage}{0.49\textwidth}
			\begin{center}
				\subfloat[(f) Texture transfer.]{\includegraphics[width=0.97\textwidth, height=0.5\textwidth]{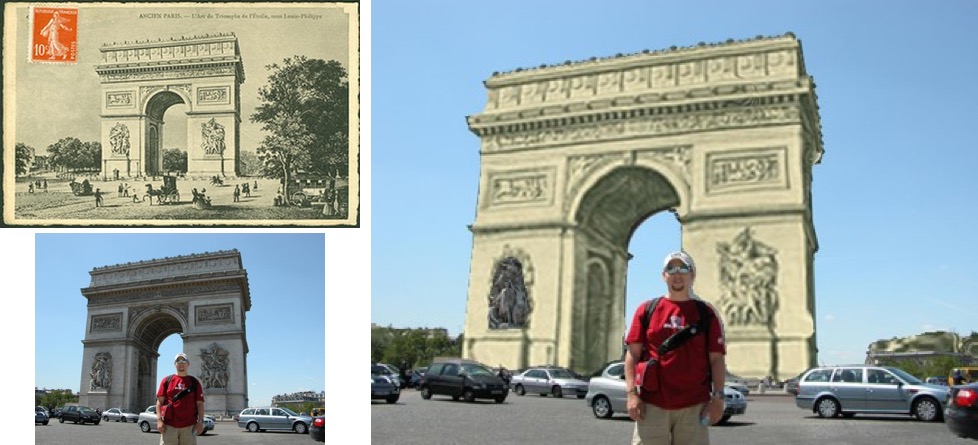}}
			\end{center}
		\end{minipage}
		\caption{RANSAC-Flow provides competitive results on a wide variety of tasks and enables new challenging applications.}
		\label{fig:teaser}
	\end{figure}

	Dense image alignment (also known as image registration) is one of the fundamental vision problems underlying many standard tasks from panorama stitching to optical flow.  Classic work on image alignment can be broadly placed into two camps: parametric and non-parametric.  Parametric methods assume that the two images are related by a global parametric transformation (e.g. affine, homography, etc), and use robust approaches, like RANSAC, to estimate this transformation.  Non-parametric methods do not make any assumptions on the type of transformation, and attempt to directly optimize some pixel agreement metric (e.g. brightness constancy constraint in optical flow and stereo).  However, both approaches have flaws: parametric methods fail (albeit gracefully) if the parametric model is only an approximation for the true transform, while non-parametric methods have trouble dealing with large displacements and large appearance changes (e.g. two photos taken at different times from different views). 
	It is natural, therefore, to consider a hybrid approach, combining the benefits of parametric and non-parametric methods together. 
	
	In this paper, we propose RANSAC-flow, a two-stage approach integrating parametric and non-parametric methods for generic dense image alignment. Figure~\ref{fig:pipline} shows an overview.  In the first stage, a classic geometry-verification method (RANSAC) is applied to a set of feature correspondences to obtain one or more candidate coarse alignments. Our method is agnostic to the particular choice of transformation(s) and features, but we've found that using multiple homographies and off-the-shelf self-supervised deep features works quite well.  In the second non-parametric stage, we refine the alignment by predicting a dense flow field for each of the candidate coarse transformations. This is achieved by self-supervised training of a deep network to optimize a standard structural similarity metric (SSIM)~\cite{wang2004} between the pixels of the warped and the original images, plus a cycle-consistency loss~\cite{zhou2016learning}. 
	
	Despite its simplicity, the proposed approach turns out to be surprisingly effective.  The coarse alignment stage takes care of large-scale viewpoint and appearance variations and, thanks to multiple homographies, is able to capture a piecewise-planar approximation of the scene structure.  The learned local flow estimation stage is able to refine the alignment to the pixel level without relying on the brightness constancy assumption. 
	As a result, our method produces competitive results across a wide range of different image alignment tasks, as shown in Figure~\ref{fig:teaser}: (a) unsupervised optical flow estimation on {\sc KITTI}~\cite{kitti2015} and {\sc Hpatches}~\cite{hpatch}, (b) visual localization on {\sc Aachen Day-Night}~\cite{sattler2018benchmarking}, (c) 2-view geometry estimation  on {\sc YFCC100M}~\cite{yfcc100m}, (d) dense image alignment, and applications to (e) detail alignment in artwork and (f) texture transfer. Our code and data are available at \url{http://imagine.enpc.fr/~shenx/RANSAC-Flow/}.

	\section{Related Work}
	
	\textbf{Feature-based image alignment.} The classic approach to align images with very different appearances is to use sparse local image features, such as SIFT~\cite{sift}, which are designed to deal with large viewpoint and illumination differences as well as clutter and occlusion. These features have to be used together with a geometric regularization step to discard false matches. This is typically done using RANSAC~\cite{fischler1981random,raguram2012usac,barath2018graph,barath2019magsac} to fit a simple geometric transformation (e.g. affine or homography)~\cite{Szeliski2006}. Recently, many works proposed to learn better local features \cite{contextdesc,superpoint,tian2017l2,mishchuk2017working,luo2018geodesc,revaud2019r2d2}. Differentiable and trainable version of RANSAC have also been developed \cite{oanet,pointnet,n3net,dfe}.
	
	Using mid-level features~\cite{singh2012unsupervised,kim2017fcss,kim2018recurrent,kim2019semantic} instead of local keypoints, proved to be beneficial for matching visual content across modalities, e.g. 3D models and paintings~\cite{aubry2014painting}.  
	Recently, \cite{artminer} learned deep mid-level features for matching across different visual media (drawings, oil paintings, frescoes, sketches, etc), and used them together with spatial verification to discover copied details in a dataset of thousands of artworks. \cite{rocco2017convolutional} used deep feature map correlations as input to a regression network on synthetic image deformations to predict the parameters of an affine or thin-plate spline deformation. Finally, transformer networks \cite{jaderberg2015spatial} can also learn parametric alignment typically as a by-product of optimizing a classification task. 
	
	\textbf{Direct image alignment.} Direct, or pixel-based, alignment has its roots in classic optical flow methods, such as Lucas-Kanade~\cite{lucas1981iterative}, who solve for a dense flow field between a pair of images under a brightness constancy assumption.  The main drawback is these methods tend to work only for very small displacements.  This has been partially addressed with hierarchical flow estimation~\cite{Szeliski2006}, as well as using local features in addition to pixels to increase robustness~\cite{brox2009large,epicflow,flowfield,cpmflow}. 
	However, all such methods are still limited to aligning very similar images, where the brightness constancy assumption mostly holds. SIFT-Flow~\cite{siftflow} was an early method that aimed at expanding optical flow-style approaches for matching pairs of images across physically distinct, and visually different scenes (and later generalized to joint image set alignment using cycle consistency~\cite{zhou2015flowweb}). Some approaches such as SCV~\cite{cech2010efficient} and MODS~\cite{mishkin2015mods}, were proposed to grow matches around initial warping. In the deep era, \cite{long2014convnets} showed that ConvNet activation features can be used for correspondence, achieving similar performance to SIFT-Flow. \cite{choy2016universal} proposed to learn matches with a Correspondence Contrastive loss, producing semi-dense matches. \cite{ncnet} introduced the idea of using 4D convolutions on the feature correlations to learn to filter neighbour consensus. 
	Note that these latter works target semantic correspondences, whereas we focus on the case when all images depict the same physical scene.
	
	\textbf{Deep Flow methods.} Deep networks can be trained to predict optical flow and to be robust to drastic appearance changes, but require adapted loss and architectures. 
	Flows can be learned in a completely supervised way using synthetic data, e.g. in~\cite{flownet,flownet2}, but transfer to real data remains a difficult problem. 
	Unsupervised training through reconstruction has been proposed in several works, targeting brightness consistency~\cite{ahmadi2016unsupervised,oaflow}, gradient consistency~\cite{ren2017unsupervised} or high SSIM~\cite{jason2016back,yin2018geonet}. 
	This idea of learning correspondences through reconstruction has been applied to video, reconstructing colors~\cite{videoCor}, predicting weights for frame reconstruction~\cite{multigrid,corrFlow}, or directly optimizing feature consistency in the warped images~\cite{timeCycle}.
	Several papers have introduced cycle consistency as an additional supervisory signal for image alignment~\cite{zhou2016learning,timeCycle}.
	Recently, feature correlation became a key part of several architectures~\cite{flownet2,pwcnet} aiming at predicting dense flows.
	Particularly relevant to us is the approach of~\cite{melekhov2019dgc} which includes a feature correlation layer in a U-Net~\cite{ronneberger2015u} architecture to improve flow resolution. A similar approach has been used in~\cite{laskar2019geometric} which predicts dense correspondences. Recently, Glu-Net~\cite{GLUNet_Truong_2020} learns dense correspondences by investigating the combined use of global and local correlation layers.
	
	\textbf{Hybrid parametric/non-parametric image alignment.}
	Classic ``plane + parallax'' approaches~\cite{sawhney19943d,kumar1994direct,irani2002direct,wulff2017optical} aimed to combine parametric and non-parametric alignment by first estimating a homography (plane) and then considering the violations from that homography (parallax). Similar ideas also appeared in stereo, e.g. model-based stereo~\cite{debevec1996}.
	Recently, \cite{yin2018geonet,cao2019learning} proposed to learn optical flow by jointly optimizing with depth and ego-motion for stereo videos. Our RANSAC-Flow is also related to the methods designed for geometric multi-model fitting, such as RPA~\cite{magri2014t}, T-linkage~\cite{magri2017multiple} and Progressive-X~\cite{barath2019progressive}.
	
	\section{Method}

	Our two-stage RANSAC-Flow method is illustrated in Figure~\ref{fig:pipline}. In this section, we describe the coarse alignment stage, the fine alignment stage, and how they can be iterated to use multiple homographies.
	
	\subsection{Coarse alignment by feature-based RANSAC}
	
	Our coarse parametric alignment is performed using RANSAC to fit a homography on a set of candidate sparse correspondences between the source and target images.  We use off-the-shelf deep features (conv4 layer of a ResNet-50 network) to obtain these correspondences. We experimented with both pre-trained ImageNet features as well as features learned via MoCo self-supervision~\cite{he2019momentum}, and obtained similar results. 
	We found it was crucial to perform feature matching at different scales. We fixed the aspect ratio of each image and extracted features at seven scales: 0.5, 0.6, 0.88, 1, 1.33, 1.66 and 2. Matches that were not symmetrically consistent were discarded. 
	The estimated homography is applied to the source image and the result is given together with the target image as input to our fine alignment. We report coarse-only baselines in Experiments section for both features as ``{\it ImageNet~\cite{resnet}+H} " and ``{\it MoCo~\cite{he2019momentum}+H} ".
	
	\subsection{Fine alignment by local flow prediction}
	\label{sec:fine}
	Given a source image $\Is$ and a target image $\It$ which have already been coarsely aligned, we want to predict a fine flow $\mathit{\flowIm_{s \to t}}$ between them. We write $\mathit{\flow_{s \to t}}$ as the mapping function associated to the flow $\mathit{\flowIm_{s \to t}}$. Since we only expect the fine alignment to work in image regions where the homography is a good approximation of the deformation, we predict a matchability mask $\mathit{\match_{s \to t}}$, indicating which correspondences are valid. In the following, we first present our objective function, then how and why we optimize it using a self-supervised deep network.
	
	\subsubsection{Objective function.}
	\label{subsec:loss}
	
	Our goal is to find a flow that warps the source into an image similar to the target. We formalize this by writing an objective function composed of three parts: a reconstruction loss $\losssim$, a matchability loss $\lossmatch$ and a cycle-consistency loss $\losscycle$. Given the pair of images($\Is$, $\It$) the total loss is:

	\begin{equation}
	\losstotal (\Is, \It) = \losssim (\Is, \It) + \matcheta \lossmatch (\Is, \It) + \flowtheta \losscycle (\Is, \It)
	\label{eq:totalloss}
	\end{equation}
	
	$\noindent$with $\matcheta$ and $\flowtheta$ hyper-parameters weighting the contribution of the matchability and cycle loss.
	We detail these three components in the following paragraphs.
	Each loss is defined pixel-wise. 
	
	\paragraph{Matchability loss.} Our matchability mask can be seen as pixel-wise weights for the reconstruction and cycle-consistency losses. These losses will thus encourage the matchability to be zero. To counteract this effect, the matchability loss encourages the matchability mask to be close to one. Since the matchabiliy should be consistent between images, we define the cycle-consistent matchability at position (x,y) in $\It$, (x',y') in $\Is$ with $(x,y) = \flow_{s \to t} (x',y')$ as: 
	\begin{equation}
	\match^{cycle}_{t}(x,y) = \match_{t \to s} (x,y) \match_{s \to t} (x',y') 
	\end{equation}
	where $\match_{s \to t} $ is the matchability predicted from source to target and $\match_{t \to s} $ the one predicted from target to source.  
	$\match^{cycle}_{t}$ will be high only if both the matchability of the corresponding pixels in the source and target are high. 
	The matchability loss encourages this cycle-consistent matchability to be close to 1: 
	
	\begin{equation}
	\label{eqn:lm}
	\lossmatch (\Is, \It) = \sum_{(x,y) \in \It} \mathopen| \match^{cycle}_{t} (x, y)- 1 \mathclose|
	\end{equation}
	Note that directly encouraging the matchability to be 1 leads to similar quantitative results, but using the cycle consistent matchability helps to identify regions that are not matchable in the qualitative results. 
	
	\paragraph{Reconstruction loss.} Reconstruction is the main term of our objective and is based on the idea that the source image warped with the predicted flow $\mathit{\flowIm_{s \to t}}$ should be aligned to the target image $\It$.
	We use the structural similarity (SSIM)~\cite{wang2004} as a robust similarity measure:
	\begin{equation}
	\losssim^{SSIM} (\Is, \It) = \sum_{(x,y) \in \It} \match^{cycle}_{t}(x,y) \left(1 - \mathit{SSIM}\left( \Is(x',y'), \It (x, y)\right) \right)
	\end{equation}

	\paragraph{Cycle consistency loss.} We enforce cycle consistency of the flow for 2-cycles:
	\begin{equation}
	\label{eqn:lf}
	\losscycle (\Is, \It)  =\sum_{(x,y) \in \It} \match^{cycle}_{t} (x,y)  \|(x',y'), \flow_{t \to s} (x, y)\|_2 
	\end{equation}

	\subsubsection{Optimization with self-supervised network.}
	Optimizing objective functions similar to the one described above is common to most optical flow approaches. However, this is known to be an extremely difficult task because of the highly non-convex nature of the objective which typically has many bad local minima. 
	Recent works on the priors implicit within deep neural network architectures~\cite{shocher2018zero,ulyanov2018deep} suggest that optimizing the flow as the output of a neural network might overcome these problems.  
	Unfortunately, our objective is still too complex to obtain good result from optimization on just a single image pair. We thus built a larger database of image pairs on which we optimize the neural network parameters in a self-supervised way (i.e. without need for any annotations). The network could then be fine-tuned on the test image pair itself, but we have found that this single-pair optimization leads to unstable results. However, if several pairs similar to the test pair are available (i.e. we have access to the entire test set), the network can be fine-tuned on this test set which leads to some improvement, as can be seen in our experiments where we systematically report our results with and without fine-tuning.
	
	To collect image pairs for the network training, we simply sample pairs of images representing the same scene and applied our coarse matching procedure. If it led to enough inliers, we added the pair to our training image set, if not we discarded it. For all the experiments, we sampled image pairs from the MegaDepth~\cite{megaDepth} scenes, using $20,000$ image pairs from $100$ scenes for training and  $500$ pairs from $30$ different scenes for validation. 
	
	
	\subsection{Multiple Homographies}
	
	The overall procedure described so far provides good results on image pairs where a single homography serves as a good (if not perfect) approximation of the overall transformation (e.g. planar scenes).  This is, however, not the case for many image pairs with strong 3D effects or large objects displacements. To address this, we iterate our alignment algorithm to let it discover more homography candidates. At each iteration, we remove feature correspondences that were inliers for the previous homographies as well as from locations inside the previously predicted matchability masks, and recompute RANSAC again.  We stop the procedure when not enough candidate correspondences remain. The full resulting flow is obtained by simply aggregating the estimated flows from each iteration together.   The number of homographies considered  depends on the input image pairs. For example, the average number of homographies we obtain from pairs for two-view geometry estimation in the YFCC100M~\cite{yfcc100m} dataset is about five.  While more complex combinations could be considered, this simple approach provides surprisingly robust results.  In our experiments, we quantitatively validate the benefits of using these multiple homographies (``{\it multi-H}").

	\subsection{Architecture and Implementation Details}
	
	\label{subsec:pipeline}
	In our fine-alignment network, the input source and target images ($\Is$, $\It$) are first processed separately by a fully-convolutional {\it feature extractor} which outputs two feature maps ($\sourceFeat$, $\targetFeat$). 
	Each feature from the source image is then compared to features in a $(2K+1)\times(2K+1)$ square neighbourhood in the target image using cosine similarity, similar to~\cite{flownet,flownet2}. This results in a $\spatialW\times\spatialH\times(2K+1)^2$ similarity tensor $\simi$ defined by:
	\begin{equation}
	\label{eqn:featmatch}
	\simi(i, j, (m+K+1) (n+K+1)) = \dfrac{\sourceFeat (i,j) . \targetFeat (i - m, j - n)}{\| \sourceFeat (i,j) \| \| \targetFeat (i - m, j - n)\|}
	\end{equation}
	where $m, n\in[-K, ..., K]$ and ``." denotes dot product. In all our experiments, we used $\kernel=3$.
	This similarity tensor is taken as input by two fully-convolutional {\it prediction networks} which predict flow and matchability. 
	
	Our {\it feature extractor} 
	is similar to the \emph{Conv3} feature extractor in ResNet-18~\cite{resnet} but with minor modifications: the first $7\times 7$ convolutional kernel of the network is replaced by a $3\times 3$ kernel without stride and all the max-poolings and strided-convolution are replaced by their anti-aliasing versions proposed in~\cite{antialiased}. These changes aim at reducing the loss of spatial resolution in the network, the output feature map being 1/8th of the resolution of the input images. The flow and matchability {\it prediction networks} are fully convolutional networks composed of three Conv+Relu+BN blocks (Convolution, Relu activation and Batch Normalization~\cite{bn}) with 512, 256, 128 filters respectively and a final convolutional layer. The  output flows and matchability are bilinearly upsampled to the resolution of the input images. Note we tried using up-convolutions, but this slightly decreased the performance while increasing the memory footprint.
	
	We use Kornia~\cite{eriba2019kornia} for homography warping. All images were resized so that their minimum dimension is 480 pixels. 
	The hyper-parameters of our objective are set to $\matcheta$ = 0.01, $\flowtheta$ = 1. We provide a study of $\matcheta$ and $\flowtheta$ in Section~\ref{sec:app_param}. The entire fine alignment model is learned from random initialization using the Adam optimizer~\cite{adam} with a learning rate of 2e-4 and momentum terms $\beta_1$, $\beta_2$  set to 0.5, 0.999. We trained only with $\losssim$ for the first 150 epochs then added $\losscycle$ for another 50 epochs and finally trained with all the losses (Equation~\ref{eq:totalloss}) for the final 50 epochs. We use a mini-batch size of 16 for all the experiments. The whole training converged in approximately 30 hours using a single GPU Geforce GTX 1080 Ti for the 20k image pairs from the MegaDepth. 
	For fine-tuning on the target dataset, we used a learning rate of 2e-4 for another 10K iterations.
	
	\captionsetup[subfloat]{labelformat=empty,farskip=2pt,captionskip=1pt}
	
	\begin{figure}[t]
		\centering     
		\subfloat[(a) Input]{\includegraphics[width=0.245\textwidth, height=0.08\textwidth]{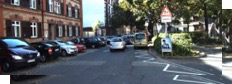}}\hfill
		\subfloat[(b) Predicted]{\includegraphics[width=0.245\textwidth, height=0.08\textwidth]{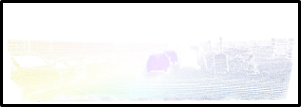}}\hfill
		\subfloat[(c) Ground truth ]{\includegraphics[width=0.245\textwidth, height=0.08\textwidth]{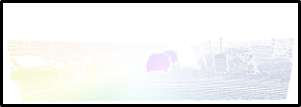}}\hfill
		\subfloat[(d) Error map]{\includegraphics[width=0.245\textwidth, height=0.08\textwidth]{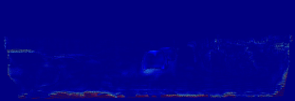}}\hfill
		\caption{Visual results on KITTI \cite{kitti2015}. We show the predicted flow, ground-truth flow and the error map in (b), (c) and (d) respectively. }
		\label{fig:kitti}
	\end{figure}

	\captionsetup[table]{farskip=2pt,captionskip=1pt,aboveskip=4pt}
	\begin{table}[t]
		\centering
		\setlength\extrarowheight{-9pt}
		\caption{(a) Dense correspondences evaluation on KITTI 2015~\cite{kitti2015} and Hpatches~\cite{hpatch}. We report the AEE (Average Endpoint Error) and Fl-all (Ratio of pixels where flow estimate is wrong by both 3 pixels and $\ge$ 5\%). The computational time for EpicFlow and FlowField is 16s and 23s respectively, while our approach takes 4s. 
			(b) Sparse correspondences evaluation on RobotCar~\cite{robotcar,crossSeason} and MegaDepth~\cite{megaDepth}. We report the accuracy over all annotated alignments for pixel error smaller than d pixels. All the images are resized to have minimum dimension 480 pixels.} 
		\begin{minipage}{0.58\textwidth}
			\scalebox{0.6}{\subfloat[(a) Dense correspondences evaluation on KITTI 2015~\cite{kitti2015} and Hpatches~\cite{hpatch}.]{
					\begin{tabular}{c||cc|cc||ccccc}
						\multicolumn{1}{c||}{\multirow{3}{*}{Method}} &\multicolumn{4}{c||}{KITTI 2015~\cite{kitti2015}} &\multicolumn{5}{c}{Hpatches~\cite{hpatch}} \\
						& \multicolumn{2}{c|}{Train (AEE $\downarrow$)} & \multicolumn{2}{c||}{Test (Fl-all $\downarrow$)} & \multicolumn{5}{c}{Viewpoint (AEE $\downarrow$)} \\
						& noc & all & noc & all & 1 & 2 & 3 & 4 & 5 \\
						\hline 
						\multicolumn{10}{c}{\textbf{Supervised Approaches}} \\
						\multicolumn{1}{c||}{FlowNet2~\cite{flownet2,melekhov2019dgc,yin2018geonet}}& 4.93 & 10.06  & 6.94 & 10.41 &5.99 &15.55 &17.09 &22.13 &30.68 \\
						
						\multicolumn{1}{c||}{PWC-Net~\cite{pwcnet,melekhov2019dgc}}& - &10.35 & \bf 6.12 &\bf 9.60 &4.43 &11.44& 15.47& 20.17& 28.30 \\
						\multicolumn{1}{c||}{Rocco~\cite{rocco2017convolutional,melekhov2019dgc}}& - &- & - &- & 9.59 &18.55 &21.15 &27.83 &35.19 \\
						\multicolumn{1}{c||}{DGC-Net~\cite{melekhov2019dgc}}& - &- & - &- & 1.55 &5.53 &8.98 &11.66 &16.70 \\
						\multicolumn{1}{c||}{DGC-Nc-Net~\cite{laskar2019geometric}} & - &- & - &- & 1.24 & 4.25 & 8.21 &9.71 &13.35\\
						\multicolumn{1}{c||}{Glu-Net~\cite{GLUNet_Truong_2020}} &  6.86 & 9.79 & - &- &  0.59 & 4.05 & 7.64 & 9.82 & 14.89\\
						
						\multicolumn{10}{c}{\textbf{Weakly Supervised Approaches}} \\
						\multicolumn{1}{c||}{ImageNet~\cite{resnet} + H} & 13.49 & 17.26 & - & - & 1.33 & 3.34 & 3.71 & 6.04 & 10.07 \\
						\multicolumn{1}{c||}{Cao et al.~\cite{cao2019learning}}& 4.19 & 5.13 & - & -&- &- &- &- &- \\
						\multicolumn{10}{c}{\textbf{Unsupervised Approaches}} \\
						
						\multicolumn{1}{c||}{Moco~\cite{he2019momentum} + H} & 13.86 & 17.60 & - & - & 1.47 & 2.96 & 3.43 & 7.73 & 10.53 \\
						\multicolumn{1}{c||}{DeepMatching~\cite{deepMatch,melekhov2019dgc}} & - & -& - & - & 5.84 & 4.63 & 12.43 & 12.17 & 22.55\\
						\multicolumn{1}{c||}{DSTFlow~\cite{dstflow}}&  6.96 & 16.79& - & 39 &- &- &- &- &- \\
						\multicolumn{1}{c||}{GeoNet~\cite{yin2018geonet}} & 6.77 & 10.81 & - & - &- &- &- &- &- \\
						\multicolumn{1}{c||}{EpicFlow~\cite{epicflow,yin2018geonet}}& 4.45 & \bf 9.57 & 16.69 & 26.29 &- &- &- &- &-  \\
						\multicolumn{1}{c||}{FlowField~\cite{flowfield}}&- & - & 10.98 & 19.80 &- &- &- &- &-  \\
						\hline 
						\multicolumn{10}{c}{\textbf{Moco Feature}} \\
						\multicolumn{1}{c||}{Ours} & 4.15 & 12.63 & 14.60  & 26.16 & 0.52 & 2.13 & 4.83 & 5.13 & 6.36   \\
						\multicolumn{1}{c||}{w/o fine-tuning} & 4.67 & 13.51 & - & - & 0.53 & \bf 2.04 & \bf 2.32 & 6.54 & 6.79   \\
						\multicolumn{1}{c||}{w/o Multi-H} & 7.04 & 14.02 & - & - &- &- &- &- &-  \\
						\multicolumn{10}{c}{\textbf{ImageNet Feature}} \\
						\multicolumn{1}{c||}{Ours} & \bf 3.87 & 12.48 & 14.12 & 25.76 & \bf 0.51 & 2.36 & 2.91 & \bf 4.41 & \bf 5.12    \\
						\multicolumn{1}{c||}{w/o fine-tuning} & 4.55 & 13.51 & - & - & \bf 0.51 & 2.37 & 2.64 & 4.49 & 5.16   \\
						\multicolumn{1}{c||}{w/o Multi-H} & 6.74 & 13.77 & - & - &- &- &- &- &-  \\
						\hline
			\end{tabular}}}
			\label{tab:dense}
		\end{minipage}
		\begin{minipage}{0.41\textwidth}
			\scalebox{0.62}{\subfloat[(b) Sparse correspondences evaluation on the RobotCar~\cite{robotcar,crossSeason} and MegaDepth~\cite{megaDepth}.]{
					\begin{tabular}{c||ccc||ccc}
						\multicolumn{1}{c||}{\multirow{3}{*}{Method}}& \multicolumn{3}{c||}{RobotCar~\cite{robotcar,crossSeason}} & \multicolumn{3}{c}{MegaDepth~\cite{megaDepth}} \\
						& \multicolumn{3}{c||}{Acc($\le$ d pixels $\uparrow$)} & \multicolumn{3}{c}{Acc($\le$ d pixels $\uparrow$)} \\
						& 1 & 3 & 5 & 1 & 3 & 5 \\
						\hline 
						
						ImageNet~\cite{resnet}+H  &1.03 & 8.12 & 19.21 & 3.49& 23.48&43.94  \\
						Moco~\cite{he2019momentum}+H  & 1.08&  8.77& 20.05& 3.70 & 25.12& 45.45  \\
						SIFT-Flow~\cite{siftflow}  &1.12 &8.13 &16.45 & 8.70& 12.19&13.30  \\
						NcNet~\cite{ncnet}+H &0.81 &7.13 &16.93 & 1.98& 14.47&32.80  \\
						DGC-Net~\cite{melekhov2019dgc}  &1.19 & 9.35 & 20.17 & 3.55& 20.33& 34.28 \\
						Glu-Net~\cite{GLUNet_Truong_2020}  &\bf 2.16 & \bf 16.77 & \bf 33.38 & 25.2& 51.0& 56.8 \\
						
						\hline
						\multicolumn{7}{c}{\textbf{Moco Feature}} \\
						Ours & 2.10& 16.07& 31.66 & \bf 53.47& 83.45& \bf 86.81\\
						w/o Multi-H &2.06 & 15.77&31.05 & 50.65& 78.34& 81.59\\
						w/o Fine-tuning & 2.09 & 15.94& 31.61 & 52.60 & \bf 83.46 & 86.80\\
						\hline
						\multicolumn{7}{c}{\textbf{ImageNet Feature}} \\
						
						Ours & 2.10& 16.09& 31.80 & 53.15& 83.34& 86.74\\
						w/o Multi-H &2.06 & 15.84&31.30 & 50.08& 77.84& 81.08\\
						w/o Fine-tuning & 2.09 & 16.00& 31.90 & 52.80& 83.31& 86.64\\
						\hline
			\end{tabular}}}
			\label{tab:sparse}
		\end{minipage}
		
	\end{table}

	\captionsetup[subfloat]{labelformat=empty,farskip=2pt,captionskip=1pt}
	
	\begin{figure}[t]
		\centering     
		\subfloat{\includegraphics[width=0.22\textwidth, height=0.14\textwidth]{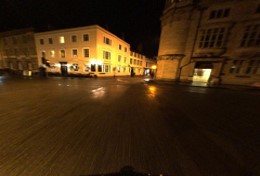}}\hfill
		\subfloat{\includegraphics[width=0.22\textwidth, height=0.14\textwidth]{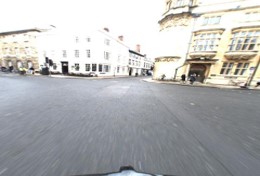}}\hfill
		\subfloat{\includegraphics[width=0.22\textwidth, height=0.14\textwidth]{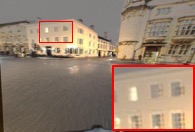}}\hfill
		\subfloat{\includegraphics[width=0.22\textwidth, height=0.14\textwidth]{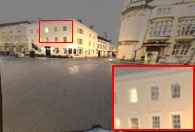}}\hfill
		\subfloat{\includegraphics[width=0.11\textwidth, height=0.14\textwidth]{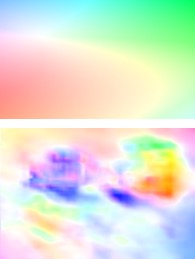}}\hfill
		
		\subfloat{\includegraphics[width=0.22\textwidth, height=0.14\textwidth]{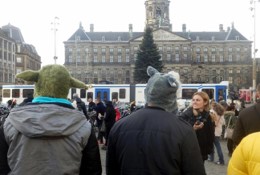}}\hfill
		\subfloat{\includegraphics[width=0.22\textwidth, height=0.14\textwidth]{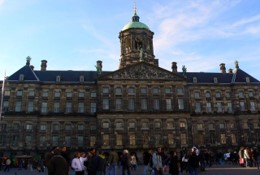}}\hfill
		\subfloat{\includegraphics[width=0.22\textwidth, height=0.14\textwidth]{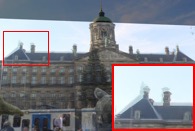}}\hfill
		\subfloat{\includegraphics[width=0.22\textwidth, height=0.14\textwidth]{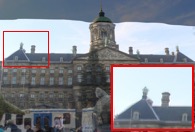}}\hfill
		\subfloat{\includegraphics[width=0.11\textwidth, height=0.14\textwidth]{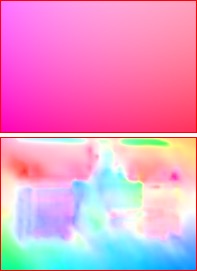}}\hfill
		
		\subfloat[(a) Source]{\includegraphics[width=0.22\textwidth, height=0.142\textwidth]{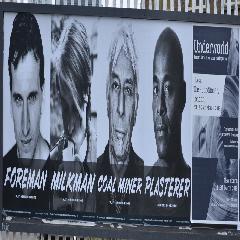}}\hfill
		\subfloat[(b) Target]{\includegraphics[width=0.22\textwidth, height=0.142\textwidth]{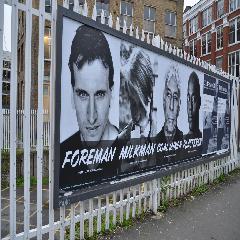}}\hfill
		\subfloat[(c) Coarse align.]{\includegraphics[width=0.22\textwidth, height=0.142\textwidth]{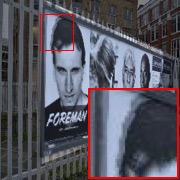}}\hfill
		\subfloat[(d) Fine align.]{\includegraphics[width=0.22\textwidth, height=0.142\textwidth]{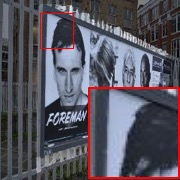}}\hfill
		\subfloat[(e) Flows]{\includegraphics[width=0.11\textwidth, height=0.142\textwidth]{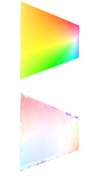}}\hfill
		\caption{Visual results on  RobotCar \cite{robotcar} (1st row), Megadepth \cite{megaDepth} (2nd row) and Hpatches \cite{hpatch} (3rd row) using one homography. We show the source and target in (a), (b). The overlapped images after coarse and fine alignment are in (c) and (d) with zoomed details. The coarse (top) and fine (bottom) flows are in (e).}
		\label{fig:vishpatches}
		
	\end{figure}

	\section{Experiments}

	In this section, we evaluate our approach in terms of resulting correspondences (Sec 4.1), downstream tasks (Sec 4.2), as well as applications to texture transfer and artwork analysis (Sec 4.3). We provide more visual results at \url{http://imagine.enpc.fr/~shenx/RANSAC-Flow/}.
	
	\subsection{Direct correspondences evaluation}
	
	\subsubsection{Optical flow.} We evaluate the quality of our dense flow on the KITTI 2015 flow~\cite{kitti2015} and Hpatches~\cite{hpatch} datasets and report the results in Table~\ref{tab:dense}. 
	
	On KITTI~\cite{kitti2015}, we evaluated both on the training and the test set since other approaches report results on one or the other. Note we could not perform an ablation study on the test set since the number of submissions to the online server is strictly limited. We report results both on non-occluded (noc) and all regions. Our results are on par with state of the art unsupervised and weakly supervised results on non-occluded regions, outperforming for example the recent approach~\cite{cao2019learning,GLUNet_Truong_2020}. Unsurprisingly, our method is much weaker on occluded regions since our algorithm is not designed specifically for optical flow performances and has no reason to handle occluded regions in a good way. We find that the largest errors are actually in occluded regions and image boundaries (Figure \ref{fig:kitti}). Interestingly, our ablations show that the multiple homographies is critical to our results even if the input images appear quite similar. 
	
	For completeness, we also present results on the Hpatches~\cite{hpatch}.  Note that Hpatches dataset is synthetically created by applying homographies to a set of real images, which would suggest that our coarse alignment alone should be enough. However, in practice, we have found that, due to the lack of feature correspondences, adding the fine flow network significantly boosts the results compared to using only our coarse approach.
	
	
	While these results show that our approach is reasonable, these datasets only contain very similar and almost aligned pairs while the main goal of our approach is to be able to handle challenging cases with strong viewpoint and appearance variations.

	\subsubsection{Sparse correspondences.} 
	Dense correspondence annotations are typically not available for extreme viewpoint and imaging condition variations. We thus evaluated our results on sparse correspondences available on the RobotCar~\cite{robotcar,crossSeason} and MegaDepth~\cite{megaDepth} datasets. In Robotcar, we evaluated on the correspondences provided by \cite{crossSeason}, which leads to approximately 340M correspondences. The task is especially challenging since the images correspond to different and challenging conditions (dawn, dusk, night, etc.) and most of the correspondences are on texture-less region such as roads where the reconstruction objective provides very little information. However, viewpoints in RobotCar are still very similar. To test our method on pairs of images with very different viewpoints, we used pairs of images from scenes of the MegaDepth~\cite{megaDepth} dataset that we didn't use for training and validation. Note that no real ground truth is available and we use as reference the result of SfM reconstructions. More precisely, we take 3D points as correspondences and randomly sample 1 600 pairs of images that shared more than 30 points, which results in approximately 367K correspondences.
	
	On both datasets, we evaluated several baselines which provide dense correspondences and were designed to handle large viewpoint changes, inluding SIFT-Flow~\cite{siftflow}, variants of NcNet~\cite{ncnet}, DGC-Net~\cite{melekhov2019dgc} and the very recent, concurrently developed Glu-Net~\cite{GLUNet_Truong_2020}. In the results provided in Table \ref{tab:sparse}, we can see that our approach is comparable to Glu-Net on RobotCar~\cite{robotcar,crossSeason} but  largely improves performances on MegeDepth~\cite{megaDepth}. We believe this is because by the large viewpoint variations on MegeDepth is better handled by our method. 
	This qualitative difference between the datasets can be seen in the visual results in Figure \ref{fig:vishpatches}. Note that we can clearly see the effect of fine flows on the zoomed details. 

	


	\subsection{Evaluation for downstream tasks.}

	\captionsetup[table]{farskip=2pt,captionskip=1pt,aboveskip=4pt}
	\begin{table}[t]
		\centering
		\setlength\extrarowheight{-9pt}
		\caption{(a) Two-view geometric estimation on YFCC100M~\cite{yfcc100m,oanet}. (b) Visual Localization on Aachen night-time~\cite{sattler2018benchmarking,sattler2012image}.} 
		\begin{minipage}{0.35\textwidth}
			\centering
			\scalebox{0.7}{\subfloat[(a) Two-view geometry, YFCC100M~\cite{yfcc100m}]{
					\begin{tabular}{c||ccc}
						Method & mAP @$5^\circ$ & mAP@$10^\circ$ &mAP@$20^\circ$
						\\
						\hline 
						\\
						SIFT~\cite{sift} & 46.83 & 68.03 & 80.58 \\
						Contextdesc~\cite{contextdesc} & 47.68 &  69.55 & \bf 84.30 \\
						Superpoint~\cite{superpoint} & 30.50 & 50.83 & 67.85 \\
						PointCN~\cite{yfcctest,oanet} & 47.98 & - & - \\
						PointNet++~\cite{pointnet++,oanet} & 46.23 & - & - \\
						$N^3$Net~\cite{n3net,oanet} & 49.13 & - & - \\
						DFE~\cite{dfe,oanet} & 49.45 & - & - \\
						OANet~\cite{oanet} & 52.18&  -& - \\
						
						\hline
						\multicolumn{4}{c}{\textbf{Moco Feature}} \\
						Ours &\bf 64.88  & \bf 73.31 & 81.56 \\
						w/o multi-H & 61.10 & 70.50 & 79.24  \\
						w/o fine-tuning & 63.48 & 72.93 & 81.59 \\
						\hline
						\multicolumn{4}{c}{\textbf{ImageNet Feature}} \\
						Ous & 62.45&  70.84& 78.99 \\
						w/o multi-H & 59.90 & 68.8 & 77.31  \\
						w/o fine-tuning & 62.10 & 70.78 & 79.07 \\
						\hline
			\end{tabular}}}
		\end{minipage}\hfill
		\begin{minipage}{0.60\textwidth}
			\centering
			\scalebox{0.7}{\subfloat[(b) Localization, Aachen night-time~\cite{sattler2018benchmarking,sattler2012image}]{
					\begin{tabular}{c||c c c}
						Method & 0.5m,$2^\circ$ & 1m,$5^\circ$ & 5m,$10^\circ$\\
						\hline
						Upright RootSIFT~\cite{sift} & 36.7& 54.1& 72.5\\
						DenseSfM~\cite{sattler2018benchmarking} & 39.8& 60.2& 84.7\\
						HAN + HN++~\cite{mishchuk2017working,mishkin2018repeatability} &39.8 &61.2 &77.6 \\
						Superpoint~\cite{superpoint} & 42.8 & 57.1 & 75.5 \\
						DELF~\cite{noh2017large} & 39.8 & 61.2 & 85.7 \\
						D2-net \cite{dusmanu2019d2} & 44.9 & 66.3 & \textbf{88.8} \\
						R2D2 \cite{revaud2019r2d2}& \bf 45.9 & 66.3 & \textbf{88.8} \\
						\hline
						\multicolumn{4}{c}{\textbf{Moco Feature}} \\
						Ours & 44.9 & \bf 68.4 & \bf 88.8   \\
						w/o Multi-H & 42.9 & \bf 68.4 & \bf 88.8  \\
						w/o Fine-tuning & 41.8 & \bf 68.4 &\bf  \bf 88.8 \\
						\hline
						\multicolumn{4}{c}{\textbf{ImageNet Feature}} \\
						Ous & 44.9& \bf 68.4& \bf 88.8 \\
						w/o Multi-H & 43.9 & 66.3 & \bf 88.8  \\
						w/o Fine-tuning &  44.9 &\bf  68.4  &\bf  88.8 \\
						
						\hline
			\end{tabular}}}
		\end{minipage}
		\label{tab:yfcc}
	\end{table}
	
	Given the limitations of the correspondence benchmarks discussed in the previous paragraph, and to demonstrate the practical interest of our results, we now evaluate our correspondences on two standard geometry estimation benchmarks where many results from competing approaches exist. Note that competing approaches typically use only sparse matches for these tasks, and being able to perform them using dense correspondences is a demonstration of the strength and originality of our method.

	\captionsetup[subfloat]{labelformat=empty,farskip=2pt,captionskip=1pt}
	\begin{figure}[t]
		\centering     
		\subfloat{\includegraphics[width=0.3\textwidth, height=0.17\textwidth]{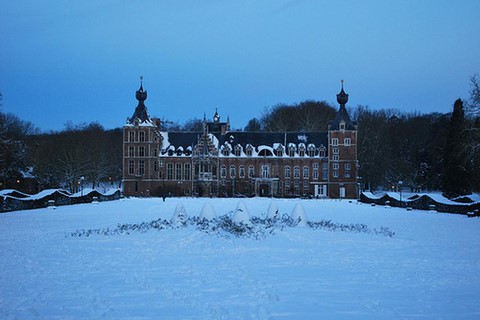}}\hfill
		\subfloat{\includegraphics[width=0.3\textwidth, height=0.17\textwidth]{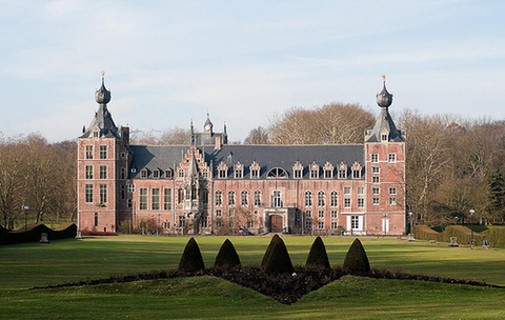}}\hfill
		\subfloat{\includegraphics[width=0.3\textwidth, height=0.17\textwidth]{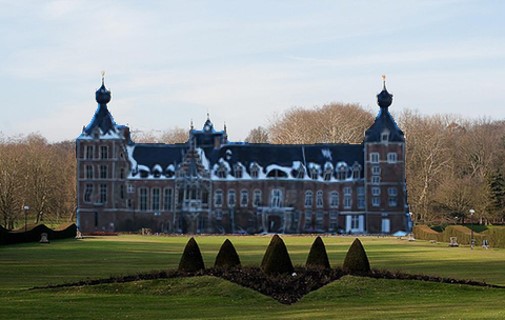}}\hfill
		\subfloat[(a) Source]{\includegraphics[width=0.3\textwidth, height=0.17\textwidth]{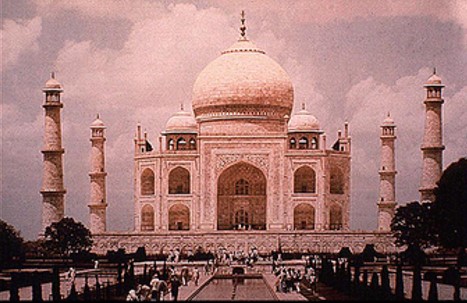}}\hfill
		\subfloat[(b) Target]{\includegraphics[width=0.3\textwidth, height=0.17\textwidth]{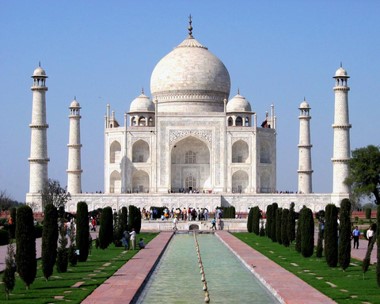}}\hfill
		\subfloat[(c) Texture transfer]{\includegraphics[width=0.3\textwidth, height=0.17\textwidth]{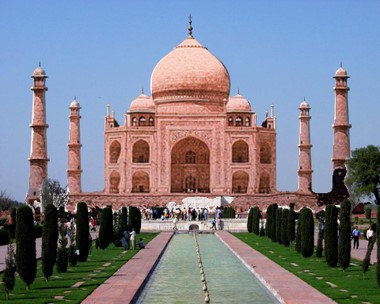}}\hfill
		
		\caption{Texture transfer : (a) source, (b) target and (c) texture transferred result.}
		\label{fig:texture_transfer}
	\end{figure}
	
	\captionsetup[subfloat]{labelformat=empty,farskip=2pt,captionskip=1pt}
	\begin{figure}[t]
		\centering     
		\subfloat[(a) Source]{\includegraphics[width=0.20\textwidth, height=0.20\textwidth]{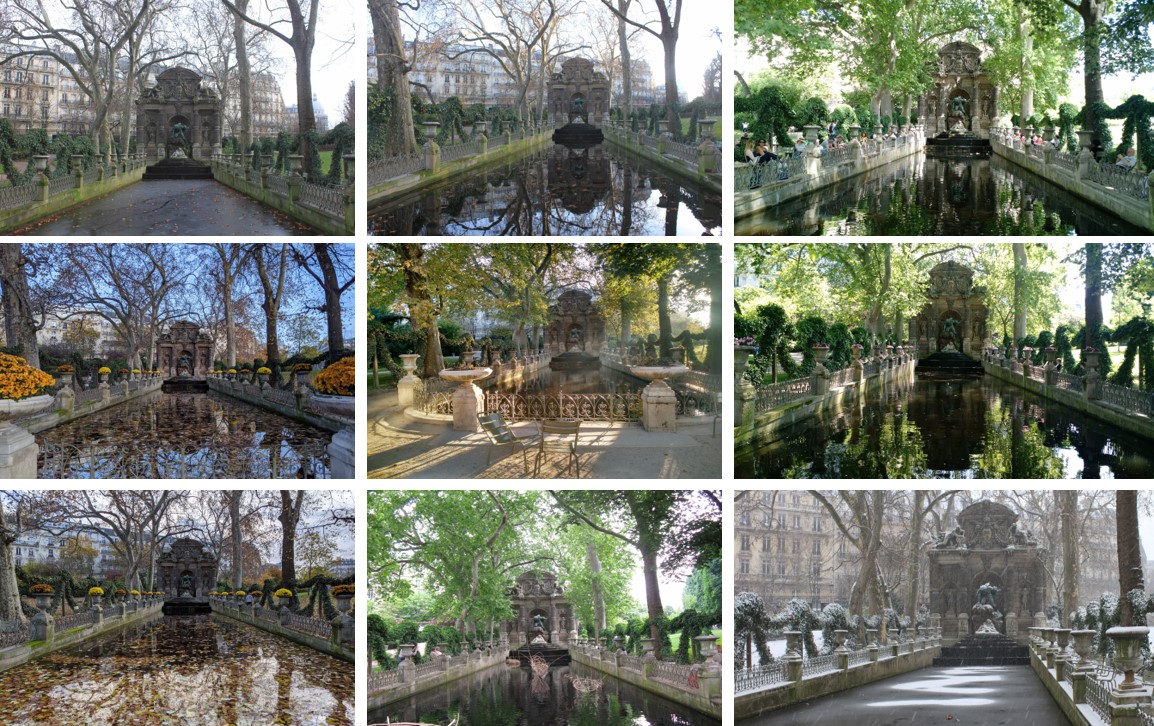}}\hfill
		\subfloat[(b) Coarse align.]{\includegraphics[width=0.26\textwidth, height=0.20\textwidth]{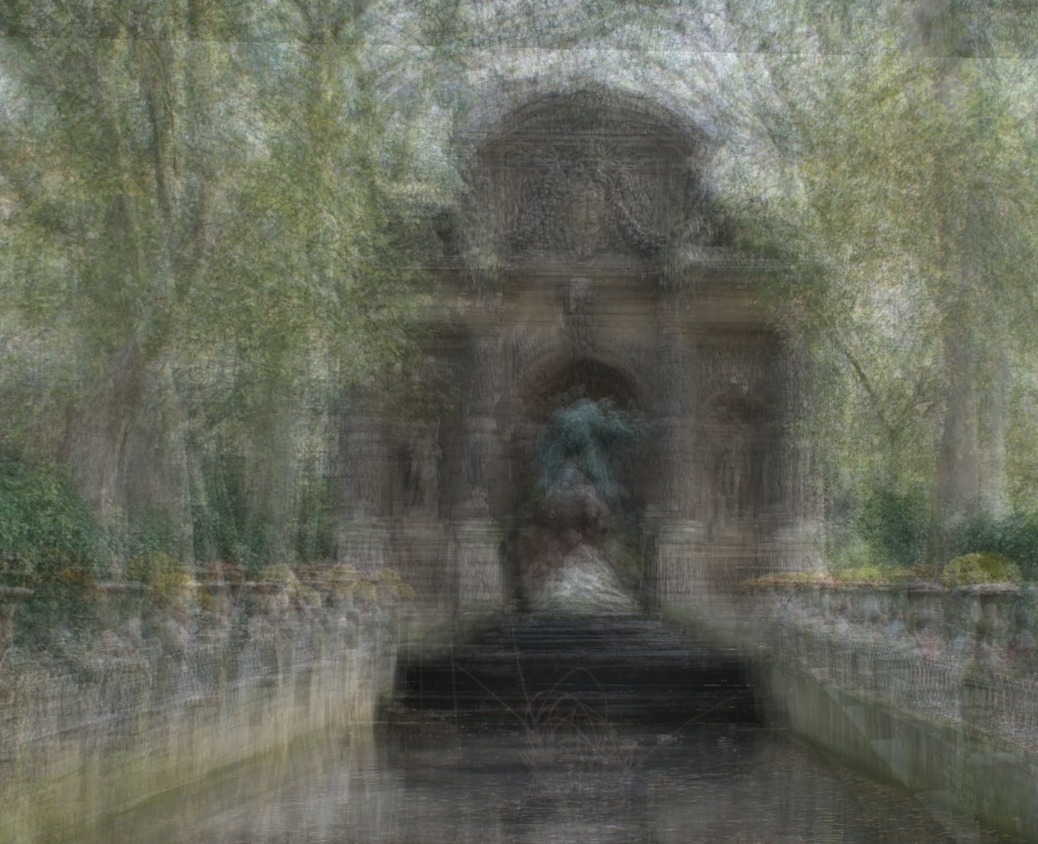}}\hfill
		\subfloat[(c) Fine align.]{\includegraphics[width=0.26\textwidth, height=0.20\textwidth]{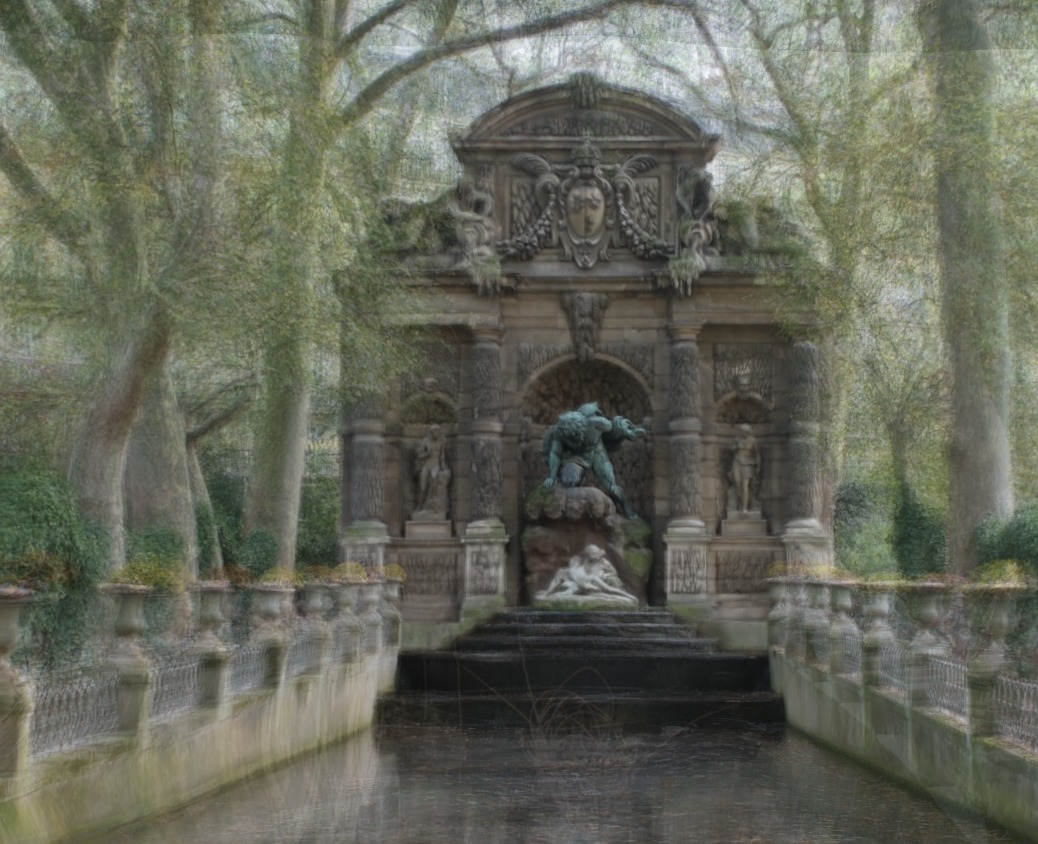}}\hfill
		\subfloat[(d) Animation]{\animategraphics[autoplay,loop, width=0.26\textwidth, height=0.20\textwidth] {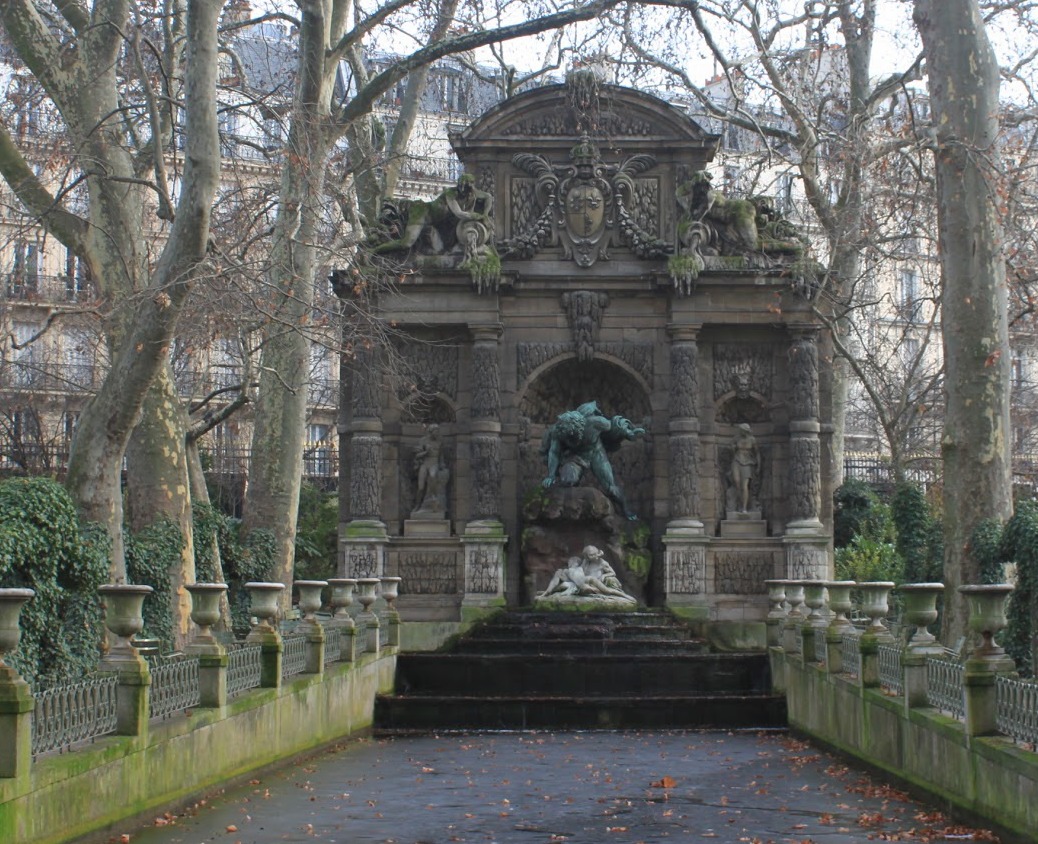}{./fig/fountain/fine/}{1}{9}}
		
		\caption{Aligning a group of Internet images from the Medici Fountain, similar to~\cite{shrivastava2011data}. We show the source images (a), the average image after coarse (b) and fine alignment (c). The animation (view with Acrobat Reader) is in (d).}
		\label{fig:align_fountain}
	\end{figure}

	\subsubsection{Two-view geometry estimation.} 
	Given a pair of views of the same scene, two-view geometry estimation aims at recovering their relative pose. To validate our approach, we follow the standard setup of~\cite{oanet} evaluating on $4\times1000$ image pairs for 4 scenes from YFCC100M~\cite{yfcc100m} dataset and reporting mAP for different thresholds on the angular differences between ground truth and predicted vectors for both rotation and translation as the error metric.  For each image pair, we use the flow we predict in regions with high matchability ($>$ 0.95) to estimate an essential matrix with RANSAC and the 5-point algorithm \cite{hartley2003multiple}. To avoid correspondences in the sky, we used the pre-trained the segmentation network provided in \cite{zhou2017scene} to remove them. While this require some supervision, this is reasonable since most of the baselines we compare to have been trained in a supervised way. 
	As can be seen in Table~\ref{tab:yfcc}, our method outperforms all the baselines by a large margin including the recent OANet~\cite{oanet} method which is trained with ground truth calibration of cameras. 
	Also note that using multiple homographies consistently boosts the performance of our method. 
	
	Once the relative pose of the cameras has been estimated, our correspondences can be used to perform stereo reconstruction from the image pair as illustrated in Figure \ref{fig:teaser}(c) and in the project webpage. Note that contrary to many stereo reconstruction methods, we can use two very different input images.

	\subsubsection{Day-Night Visual Localization.} Another task we performed is visual localization. We evaluate on the local feature challenge of the Visual Localization benchmark~\cite{sattler2018benchmarking,sattler2012image}. For each of the 98 night-time images contained in the dataset, up to 20 relevant day-time images with known camera poses are given. We followed evaluation protocol from~\cite{sattler2018benchmarking} and first compute image matching for a list of image pairs and then give them as input to COLMAP \cite{schonberger2016structure} that provides a localisation estimation for the queries. To limit the number of correspondences we use only correspondences on a sparse set of keypoints using the Superpoint~\cite{superpoint}. 
	Our results are reported in Table~\ref{tab:yfcc}(b) and are on par with state of the art results. 

	\captionsetup[subfloat]{labelformat=empty,farskip=2pt,captionskip=1pt}
	
	\begin{figure}[t]
		\centering     
		\subfloat{\includegraphics[width=0.13\textwidth]{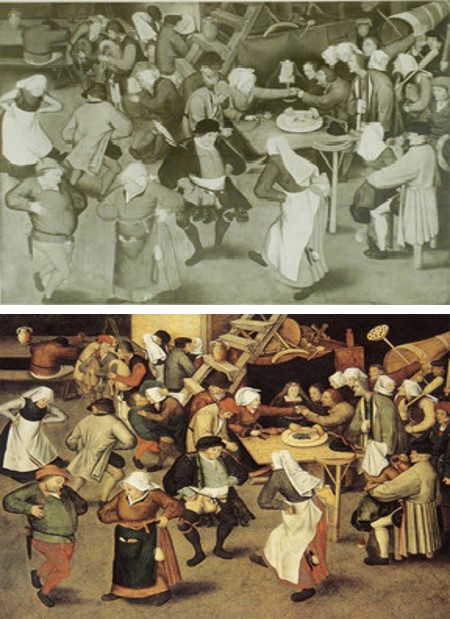}}\hfill
		\subfloat{\includegraphics[width=0.23\textwidth]{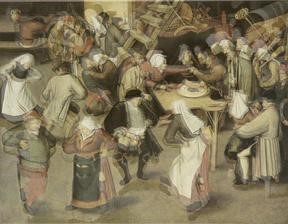}}\hfill
		\subfloat{\includegraphics[width=0.23\textwidth]{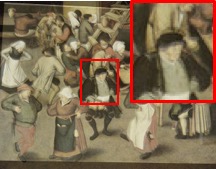}}\hfill
		\subfloat{\includegraphics[width=0.23\textwidth]{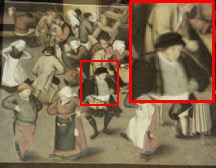}}\hfill
		\subfloat{\includegraphics[width=0.135\textwidth]{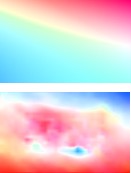}}\hfill
		\subfloat{\includegraphics[width=0.13\textwidth]{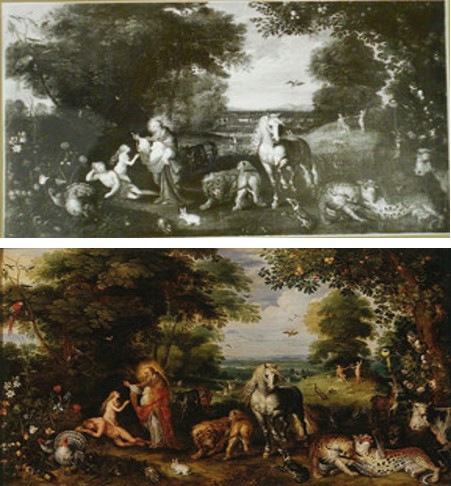}}\hfill
		\subfloat{\includegraphics[width=0.23\textwidth]{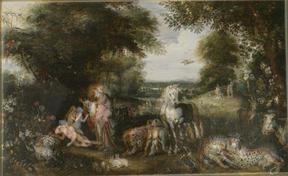}}\hfill
		\subfloat{\includegraphics[width=0.23\textwidth]{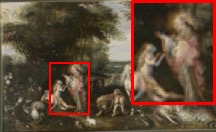}}\hfill
		\subfloat{\includegraphics[width=0.23\textwidth]{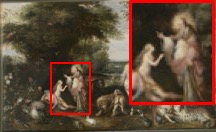}}\hfill
		\subfloat{\includegraphics[width=0.135\textwidth]{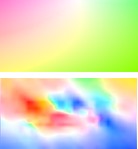}}\hfill
		\subfloat[(a) Inputs]{\includegraphics[width=0.13\textwidth]{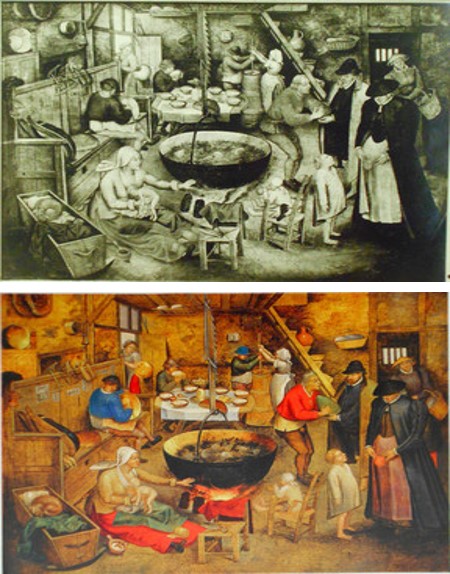}}\hfill
		\subfloat[(b) W/o align. ]{\includegraphics[width=0.23\textwidth]{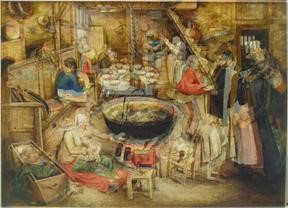}}\hfill
		\subfloat[(c) Coarse align.]{\includegraphics[width=0.23\textwidth]{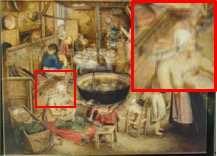}}\hfill
		\subfloat[(d) Fine align.]{\includegraphics[width=0.23\textwidth]{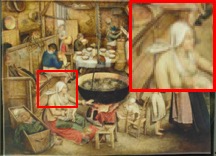}}\hfill
		\subfloat[(e) Flows]{\includegraphics[width=0.135\textwidth]{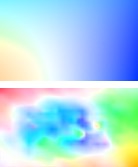}}\hfill
		
		\caption{Aligning pairs of similar artworks from the Brueghel~\cite{brueghel}: We show the pairs in (a). The average images without alignment, after coarse and fine alignment are in (b), (c) and (d). The coarse (top) and fine (bottom) flows are in (e).}
		\label{fig:brueghel_align}
	\end{figure}
	
	\captionsetup[subfloat]{labelformat=empty,farskip=2pt,captionskip=1pt}
	\begin{figure}[t]
		\centering     
		\subfloat{\includegraphics[width=0.19\textwidth, height=0.22\textwidth]{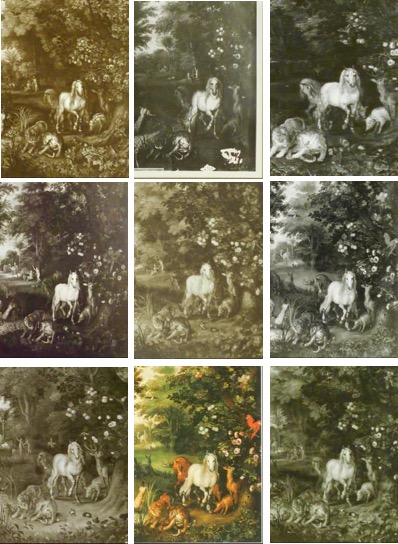}}\hfill\hfill
		\subfloat{\includegraphics[width=0.19\textwidth, height=0.22\textwidth]{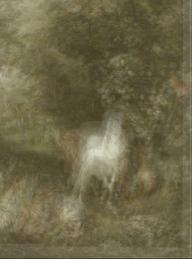}}\hfill
		\subfloat{\includegraphics[width=0.19\textwidth, height=0.22\textwidth]{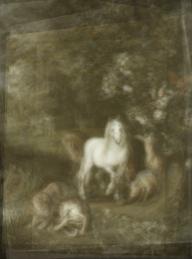}}\hfill
		\subfloat{\includegraphics[width=0.19\textwidth, height=0.22\textwidth]{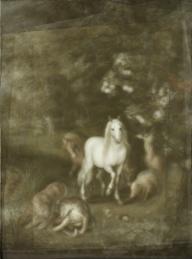}}\hfill\hfill
		\subfloat{\animategraphics[autoplay,loop, width=0.19\textwidth, height=0.22\textwidth] {1}{./fig/detail/V5/fine}{0}{8}}
		\hfill
		\subfloat{\includegraphics[width=0.19\textwidth, height=0.09\textwidth]{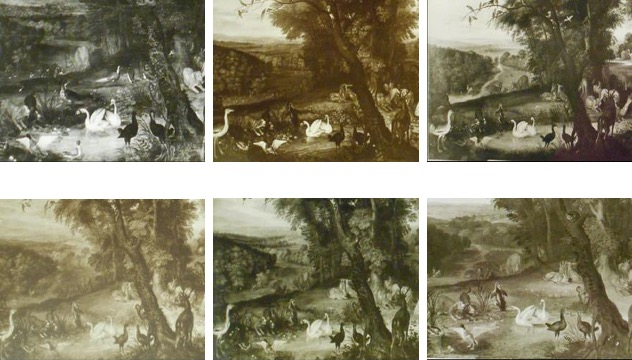}}\hfill\hfill
		\subfloat{\includegraphics[width=0.19\textwidth, height=0.09\textwidth]{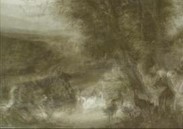}}\hfill
		\subfloat{\includegraphics[width=0.19\textwidth, height=0.09\textwidth]{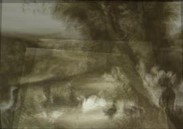}}\hfill
		\subfloat{\includegraphics[width=0.19\textwidth, height=0.09\textwidth]{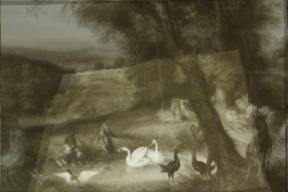}}\hfill\hfill
		\subfloat{\animategraphics[autoplay,loop, width=0.19\textwidth, height=0.09\textwidth] {1}{./fig/detail/v3/fine}{1}{6}}
		\hfill
		\subfloat[(a) Source]{\includegraphics[width=0.19\textwidth, height=0.20\textwidth]{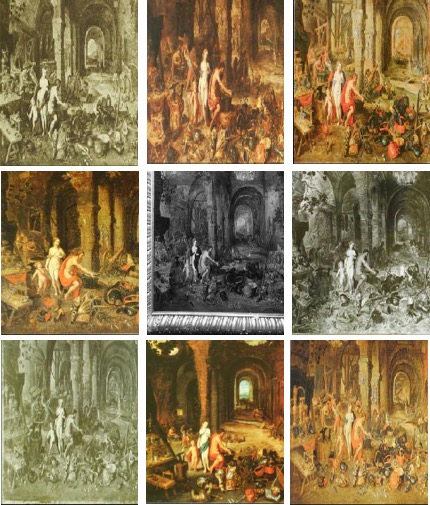}}\hfill\hfill
		\subfloat[(b) \cite{artminer}]{\includegraphics[width=0.19\textwidth, height=0.20\textwidth]{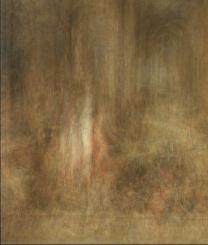}}\hfill
		\subfloat[(c) Coarse align.]{\includegraphics[width=0.19\textwidth, height=0.20\textwidth]{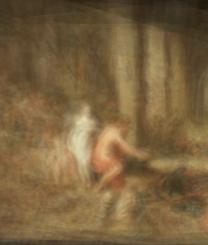}}\hfill
		\subfloat[(d) Fine align.]{\includegraphics[width=0.19\textwidth, height=0.20\textwidth]{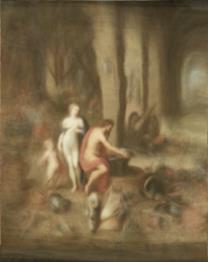}}\hfill\hfill
		\subfloat[(e) Animation]{\animategraphics[autoplay,loop, width=0.19\textwidth, height=0.20\textwidth] {1}{./fig/detail/v2/fine}{1}{21}}
		\caption{Aligning details discovered by ~\cite{artminer}: (a) sources; average from~\cite{artminer} (b), with coarse (c) and fine (d) alignment; (e) animation (view with Acrobat Reader).}
		\label{fig:align_detail}
	\end{figure}
	\begin{figure}[t]
		\begin{center}	
			\includegraphics[width=1.0\textwidth]{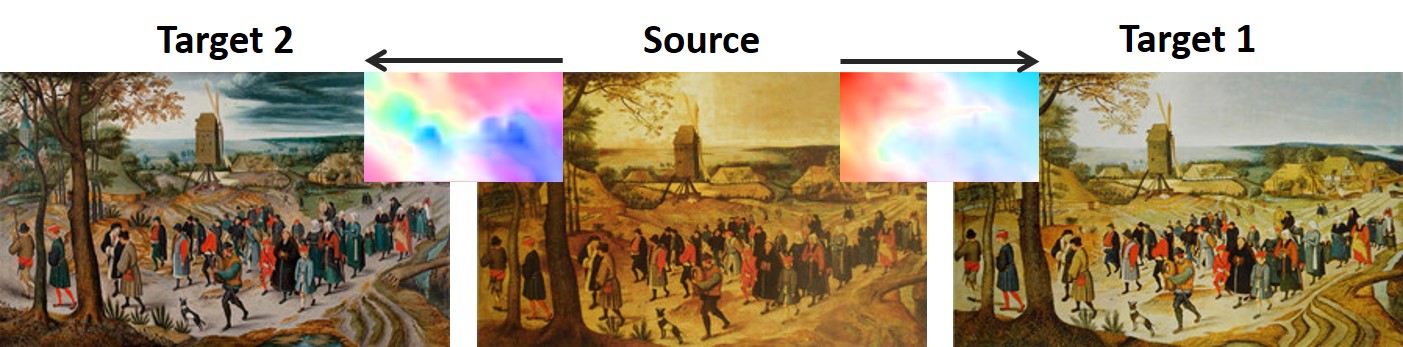}	
		\end{center}
		\caption{Analyzing copy process from flow. The flow is smooth from the middle to the right one, while it is irregular from the middle to the left one.}
		\label{fig:copy_process}
	\end{figure}

	\subsection{Applications} 
	One of the most exciting aspect of our approach is that it enables new applications based on the fine alignment of historical, internet or artistic images.
	
	{\bf Texture transfer.} Our approach can be used to transfer texture between images. In Figure~\ref{fig:texture_transfer} and~\ref{fig:teaser}(f) we show results using historical and modern images from the LTLL dataset~\cite{ltll}. We use the pre-trained segmentation network of~\cite{zhou2018semantic}, and transfer the texture from the source to the target building regions.
	
	{\bf Internet images alignment.} As visualized in Figures \ref{fig:teaser}(d) and \ref{fig:align_fountain}, we can align sets of internet images, similar to \cite{shrivastava2011data}. Even if our image set is not precisely the same, much more details can be seen in the average of our fine-aligned images.
	
	{\bf Artwork analysis.}
	Finding and matching near-duplicate patterns is an important problem for art historians. 
	Computationally, it is difficult because the duplicate appearance can be very different~\cite{artminer}.
	In Figure~\ref{fig:brueghel_align}, we show visual results of aligning different versions of artworks from the Brueghel dataset~\cite{artminer} with our coarse and fine alignment. 
	We can clearly see that a simple homography is not sufficient and that the fine alignment improves results by identifying complex displacements. 
	The fine flow can thus be used to provide insights on Brueghel's copy process. Indeed, we found that some artworks were copied in a spatially consistent way, while in others, different parts of the picture were not aligned with each other. This can be clearly seen in the flows in Figure~\ref{fig:copy_process}, which are either very regular or very discontinuous.
	The same process can be applied to more than a single pair of images, as illustrated in Figure~\ref{fig:teaser}(e) and~\ref{fig:align_detail} where we align together many similar details identified by~\cite{artminer}. Visualizing the succession of the finely aligned images allows to identify their differences.

	\section{Conclusion}
	
	We have introduced a new unsupervised method for generic dense image alignment which performs well on a wide range of tasks. Our main insight is to combine the advantages of parametric and non-parametric methods in a two-stage approach and to use multiple homography estimations as initializations for fine flow prediction. We also demonstrated it allows new applications for artwork analysis.
	
	\paragraph{Acknowledgements:}
	This work was supported by ANR project EnHerit ANR-17-CE23-0008, project Rapid Tabasco, NSF IIS-1633310, grants from SAP and Berkeley CLTC, and gifts from Adobe. We thank Shiry Ginosar, Thibault Groueix and Michal Irani for helpful discussions, and Elizabeth Alice Honig for her help in building the Brueghel dataset.
	
	\clearpage
	%
	%
	\bibliographystyle{splncs04}
	\bibliography{arXiv}

\begin{thebibliography}{10}
\providecommand{\url}[1]{\texttt{#1}}
\providecommand{\urlprefix}{URL }
\providecommand{\doi}[1]{https://doi.org/#1}

\bibitem{brueghel}
Brueghel family: Jan brueghel the elder." the brueghel family database.
  university of california, berkeley. \url{http://www.janbrueghel.net/},
  accessed: 2018-10-16

\bibitem{ahmadi2016unsupervised}
Ahmadi, A., Patras, I.: Unsupervised convolutional neural networks for motion
  estimation. In: International Conference on Image Processing (2016)

\bibitem{aubry2014painting}
Aubry, M., Russell, B.C., Sivic, J.: Painting-to-3d model alignment via
  discriminative visual elements. ACM Transactions on Graphics (ToG)  (2014)

\bibitem{flowfield}
Bailer, C., Taetz, B., Stricker, D.: Flow fields: Dense correspondence fields
  for highly accurate large displacement optical flow estimation. In:
  Proceedings of the IEEE International Conference on Computer Vision (2015)

\bibitem{hpatch}
Balntas, V., Lenc, K., Vedaldi, A., Mikolajczyk, K.: Hpatches: A benchmark and
  evaluation of handcrafted and learned local descriptors. In: Proceedings of
  the IEEE Conference on Computer Vision and Pattern Recognition (2017)

\bibitem{barath2018graph}
Barath, D., Matas, J.: Graph-cut ransac. In: Proceedings of the IEEE Conference
  on Computer Vision and Pattern Recognition (2018)

\bibitem{barath2019progressive}
Barath, D., Matas, J.: Progressive-x: Efficient, anytime, multi-model fitting
  algorithm. In: Proceedings of the IEEE International Conference on Computer
  Vision (2019)

\bibitem{barath2019magsac}
Barath, D., Matas, J., Noskova, J.: Magsac: marginalizing sample consensus. In:
  Proceedings of the IEEE Conference on Computer Vision and Pattern Recognition
  (2019)

\bibitem{brox2009large}
Brox, T., Bregler, C., Malik, J.: Large displacement optical flow. In:
  Proceedings of the IEEE Conference on Computer Vision and Pattern Recognition
  (2009)

\bibitem{cao2019learning}
Cao, Z., Kar, A., Hane, C., Malik, J.: Learning independent object motion from
  unlabelled stereoscopic videos. In: Proceedings of the IEEE Conference on
  Computer Vision and Pattern Recognition (2019)

\bibitem{cech2010efficient}
Cech, J., Matas, J., Perdoch, M.: Efficient sequential correspondence selection
  by cosegmentation. IEEE Transactions on Pattern Analysis and Machine
  Intelligence  (2010)

\bibitem{choy2016universal}
Choy, C.B., Gwak, J., Savarese, S., Chandraker, M.: Universal correspondence
  network. In: Advances in Neural Information Processing Systems (2016)

\bibitem{debevec1996}
Debevec, P.E., Taylor, C.J., Malik, J.: Modeling and rendering architecture
  from photographs: A hybrid geometry-and image-based approach. In: Proceedings
  of the 23rd annual conference on Computer graphics and interactive techniques
  (1996)

\bibitem{superpoint}
DeTone, D., Malisiewicz, T., Rabinovich, A.: Superpoint: Self-supervised
  interest point detection and description. In: Proceedings of the IEEE
  Conference on Computer Vision and Pattern Recognition Workshops (2018)

\bibitem{flownet}
Dosovitskiy, A., Fischer, P., Ilg, E., Hausser, P., Hazirbas, C., Golkov, V.,
  Van Der~Smagt, P., Cremers, D., Brox, T.: Flownet: Learning optical flow with
  convolutional networks. In: Proceedings of the IEEE international conference
  on computer vision (2015)

\bibitem{dusmanu2019d2}
Dusmanu, M., Rocco, I., Pajdla, T., Pollefeys, M., Sivic, J., Torii, A.,
  Sattler, T.: D2-net: A trainable cnn for joint description and detection of
  local features. In: Proceedings of the IEEE Conference on Computer Vision and
  Pattern Recognition (2019)

\bibitem{ltll}
Fernando, B., Tommasi, T., Tuytelaars, T.: Location recognition over large time
  lags. Computer Vision and Image Understanding  (2015)

\bibitem{fischler1981random}
Fischler, M.A., Bolles, R.C.: Random sample consensus: a paradigm for model
  fitting with applications to image analysis and automated cartography.
  Communications of the ACM  (1981)

\bibitem{hartley2003multiple}
Hartley, R., Zisserman, A.: Multiple view geometry in computer vision.
  Cambridge university press (2003)

\bibitem{he2019momentum}
He, K., Fan, H., Wu, Y., Xie, S., Girshick, R.: Momentum contrast for
  unsupervised visual representation learning. In: Proceedings of the IEEE
  Conference on Computer Vision and Pattern Recognition (2020)

\bibitem{resnet}
He, K., Zhang, X., Ren, S., Sun, J.: Deep residual learning for image
  recognition. In: Proceedings of the IEEE conference on Computer Vision and
  Pattern Recognition (2016)

\bibitem{cpmflow}
Hu, Y., Song, R., Li, Y.: Efficient coarse-to-fine patchmatch for large
  displacement optical flow. In: Proceedings of the IEEE Conference on Computer
  Vision and Pattern Recognition (2016)

\bibitem{flownet2}
Ilg, E., Mayer, N., Saikia, T., Keuper, M., Dosovitskiy, A., Brox, T.: Flownet
  2.0: Evolution of optical flow estimation with deep networks. In: Proceedings
  of the IEEE Conference on Computer Vision and Pattern Recognition (2017)

\bibitem{bn}
Ioffe, S., Szegedy, C.: Batch normalization: Accelerating deep network training
  by reducing internal covariate shift. In: International Conference on Machine
  Learning (2015)

\bibitem{irani2002direct}
Irani, M., Anandan, P., Cohen, M.: Direct recovery of planar-parallax from
  multiple frames. IEEE Transactions on Pattern Analysis and Machine
  Intelligence  (2002)

\bibitem{jaderberg2015spatial}
Jaderberg, M., Simonyan, K., Zisserman, A., et~al.: Spatial transformer
  networks. In: Advances in Neural Information Processing Systems (2015)

\bibitem{jason2016back}
Jason, J.Y., Harley, A.W., Derpanis, K.G.: Back to basics: Unsupervised
  learning of optical flow via brightness constancy and motion smoothness. In:
  Proceedings of the European Conference on Computer Vision (2016)

\bibitem{kim2018recurrent}
Kim, S., Lin, S., JEON, S.R., Min, D., Sohn, K.: Recurrent transformer networks
  for semantic correspondence. In: Advances in Neural Information Processing
  Systems (2018)

\bibitem{kim2017fcss}
Kim, S., Min, D., Ham, B., Jeon, S., Lin, S., Sohn, K.: Fcss: Fully
  convolutional self-similarity for dense semantic correspondence. In:
  Proceedings of the IEEE Conference on Computer Vision and Pattern Recognition
  (2017)

\bibitem{kim2019semantic}
Kim, S., Min, D., Jeong, S., Kim, S., Jeon, S., Sohn, K.: Semantic attribute
  matching networks. In: Proceedings of the IEEE Conference on Computer Vision
  and Pattern Recognition (2019)

\bibitem{adam}
Kingma, D.P., Ba, J.: Adam: A method for stochastic optimization. In:
  International Conference for Learning Representations (2014)

\bibitem{multigrid}
Kong, S., Fowlkes, C.: Multigrid predictive filter flow for unsupervised
  learning on videos. arXiv preprint arXiv:1904.01693  (2019)

\bibitem{kumar1994direct}
Kumar, R., Anandan, P., Hanna, K.: Direct recovery of shape from multiple
  views: A parallax based approach. In: Proceedings of 12th International
  Conference on Pattern Recognition (1994)

\bibitem{corrFlow}
Lai, Z., Xie, W.: Self-supervised learning for video correspondence flow. In:
  BMVC (2019)

\bibitem{crossSeason}
Larsson, M., Stenborg, E., Hammarstrand, L., Pollefeys, M., Sattler, T., Kahl,
  F.: A cross-season correspondence dataset for robust semantic segmentation.
  In: Proceedings of the IEEE Conference on Computer Vision and Pattern
  Recognition (2019)

\bibitem{laskar2019geometric}
Laskar, Z., Melekhov, I., Tavakoli, H.R., Ylioinas, J., Kannala, J.: Geometric
  image correspondence verification by dense pixel matching. In: Winter
  Conference on Applications of Computer Vision (2020)

\bibitem{megaDepth}
Li, Z., Snavely, N.: Megadepth: Learning single-view depth prediction from
  internet photos. In: Proceedings of the IEEE Conference on Computer Vision
  and Pattern Recognition (2018)

\bibitem{siftflow}
Liu, C., Yuen, J., Torralba, A.: Sift flow: Dense correspondence across scenes
  and its applications. IEEE Transactions on Pattern Analysis and Machine
  Intelligence  (2010)

\bibitem{long2014convnets}
Long, J.L., Zhang, N., Darrell, T.: Do convnets learn correspondence? In:
  Advances in neural information processing systems (2014)

\bibitem{sift}
Lowe, D.G.: Distinctive image features from scale-invariant keypoints.
  International Journal of Computer Vision  (2004)

\bibitem{lucas1981iterative}
Lucas, B.D., Kanade, T., et~al.: An iterative image registration technique with
  an application to stereo vision  (1981)

\bibitem{contextdesc}
Luo, Z., Shen, T., Zhou, L., Zhang, J., Yao, Y., Li, S., Fang, T., Quan, L.:
  Contextdesc: Local descriptor augmentation with cross-modality context.
  Proceedings of the IEEE Conference on Computer Vision and Pattern Recognition
   (2019)

\bibitem{luo2018geodesc}
Luo, Z., Shen, T., Zhou, L., Zhu, S., Zhang, R., Yao, Y., Fang, T., Quan, L.:
  Geodesc: Learning local descriptors by integrating geometry constraints. In:
  Proceedings of the European Conference on Computer Vision (2018)

\bibitem{robotcar}
Maddern, W., Pascoe, G., Linegar, C., Newman, P.: 1 year, 1000 km: The oxford
  robotcar dataset. The International Journal of Robotics Research  (2017)

\bibitem{magri2014t}
Magri, L., Fusiello, A.: T-linkage: A continuous relaxation of j-linkage for
  multi-model fitting. In: Proceedings of the IEEE Conference on Computer
  Vision and Pattern Recognition (2014)

\bibitem{magri2017multiple}
Magri, L., Fusiello, A.: Multiple structure recovery via robust preference
  analysis. Image and Vision Computing  (2017)

\bibitem{melekhov2019dgc}
Melekhov, I., Tiulpin, A., Sattler, T., Pollefeys, M., Rahtu, E., Kannala, J.:
  Dgc-net: Dense geometric correspondence network. In: Winter Conference on
  Applications of Computer Vision (2019)

\bibitem{kitti2015}
Menze, M., Geiger, A.: Object scene flow for autonomous vehicles. In:
  Proceedings of the IEEE Conference on Computer Vision and Pattern Recognition
  (2015)

\bibitem{mishchuk2017working}
Mishchuk, A., Mishkin, D., Radenovic, F., Matas, J.: Working hard to know your
  neighbor's margins: Local descriptor learning loss. In: Advances in Neural
  Information Processing Systems (2017)

\bibitem{mishkin2015mods}
Mishkin, D., Matas, J., Perdoch, M.: Mods: Fast and robust method for two-view
  matching. Computer Vision and Image Understanding  (2015)

\bibitem{mishkin2018repeatability}
Mishkin, D., Radenovic, F., Matas, J.: Repeatability is not enough: Learning
  affine regions via discriminability. In: Proceedings of the European
  Conference on Computer Vision (2018)

\bibitem{yfcctest}
Moo~Yi, K., Trulls, E., Ono, Y., Lepetit, V., Salzmann, M., Fua, P.: Learning
  to find good correspondences. In: Proceedings of the IEEE Conference on
  Computer Vision and Pattern Recognition (2018)

\bibitem{noh2017large}
Noh, H., Araujo, A., Sim, J., Weyand, T., Han, B.: Large-scale image retrieval
  with attentive deep local features. In: Proceedings of the IEEE International
  Conference on Computer Vision (2017)

\bibitem{n3net}
Pl{\"o}tz, T., Roth, S.: Neural nearest neighbors networks. In: Advances in
  Neural Information Processing Systems (2018)

\bibitem{GLUNet_Truong_2020}
Prune, T., Martin, D., Radu, T.: {GLU-Net}: Global-local universal network for
  dense flow and correspondences. In: Proceedings of the IEEE Conference on
  Computer Vision and Pattern Recognition (2020)

\bibitem{pointnet}
Qi, C.R., Su, H., Mo, K., Guibas, L.J.: Pointnet: Deep learning on point sets
  for 3d classification and segmentation. In: Proceedings of the IEEE
  Conference on Computer Vision and Pattern Recognition (2017)

\bibitem{pointnet++}
Qi, C.R., Yi, L., Su, H., Guibas, L.J.: Pointnet++: Deep hierarchical feature
  learning on point sets in a metric space. In: Advances in Neural Information
  Processing Systems (2017)

\bibitem{raguram2012usac}
Raguram, R., Chum, O., Pollefeys, M., Matas, J., Frahm, J.M.: Usac: a universal
  framework for random sample consensus. IEEE Transactions on Pattern Analysis
  and Machine Intelligence  (2012)

\bibitem{dfe}
Ranftl, R., Koltun, V.: Deep fundamental matrix estimation. In: Proceedings of
  the European Conference on Computer Vision (2018)

\bibitem{ren2017unsupervised}
Ren, Z., Yan, J., Ni, B., Liu, B., Yang, X., Zha, H.: Unsupervised deep
  learning for optical flow estimation. In: Thirty-First AAAI Conference on
  Artificial Intelligence (2017)

\bibitem{dstflow}
Ren, Z., Yan, J., Ni, B., Liu, B., Yang, X., Zha, H.: Unsupervised deep
  learning for optical flow estimation. In: Thirty-First AAAI Conference on
  Artificial Intelligence (2017)

\bibitem{epicflow}
Revaud, J., Weinzaepfel, P., Harchaoui, Z., Schmid, C.: Epicflow:
  Edge-preserving interpolation of correspondences for optical flow. In:
  Proceedings of the IEEE Conference on Computer Vision and Pattern Recognition
  (2015)

\bibitem{deepMatch}
Revaud, J., Weinzaepfel, P., Harchaoui, Z., Schmid, C.: Deepmatching:
  Hierarchical deformable dense matching. International Journal of Computer
  Vision  (2016)

\bibitem{revaud2019r2d2}
Revaud, J., Weinzaepfel, P., de~Souza, C.R., Humenberger, M.: {R2D2:}
  repeatable and reliable detector and descriptor. In: Advances in Neural
  Information Processing Systems (2019)

\bibitem{eriba2019kornia}
Riba, E., Mishkin, D., Ponsa, D., Rublee, E., Bradski, G.: Kornia: an open
  source differentiable computer vision library for pytorch. In: Winter
  Conference on Applications of Computer Vision (2020)

\bibitem{rocco2017convolutional}
Rocco, I., Arandjelovic, R., Sivic, J.: Convolutional neural network
  architecture for geometric matching. In: Proceedings of the IEEE Conference
  on Computer Vision and Pattern Recognition (2017)

\bibitem{ncnet}
Rocco, I., Cimpoi, M., Arandjelovi{\'c}, R., Torii, A., Pajdla, T., Sivic, J.:
  Neighbourhood consensus networks. In: Advances in Neural Information
  Processing Systems (2018)

\bibitem{ronneberger2015u}
Ronneberger, O., Fischer, P., Brox, T.: U-net: Convolutional networks for
  biomedical image segmentation. In: International Conference on Medical image
  computing and computer-assisted intervention (2015)

\bibitem{sattler2018benchmarking}
Sattler, T., Maddern, W., Toft, C., Torii, A., Hammarstrand, L., Stenborg, E.,
  Safari, D., Okutomi, M., Pollefeys, M., Sivic, J., et~al.: Benchmarking 6dof
  outdoor visual localization in changing conditions. In: Proceedings of the
  IEEE Conference on Computer Vision and Pattern Recognition (2018)

\bibitem{sattler2012image}
Sattler, T., Weyand, T., Leibe, B., Kobbelt, L.: Image retrieval for
  image-based localization revisited. In: BMVC (2012)

\bibitem{sawhney19943d}
Sawhney, H.S.: 3d geometry from planar parallax. In: Proceedings of the IEEE
  Conference on Computer Vision and Pattern Recognition (1994)

\bibitem{schonberger2016structure}
Schonberger, J.L., Frahm, J.M.: Structure-from-motion revisited. In:
  Proceedings of the IEEE Conference on Computer Vision and Pattern Recognition
  (2016)

\bibitem{artminer}
Shen, X., Efros, A.A., Aubry, M.: Discovering visual patterns in art
  collections with spatially-consistent feature learning. In: Proceedings IEEE
  Conf. on Computer Vision and Pattern Recognition (2019)

\bibitem{shocher2018zero}
Shocher, A., Cohen, N., Irani, M.: “zero-shot” super-resolution using deep
  internal learning. In: Proceedings of the IEEE Conference on Computer Vision
  and Pattern Recognition (2018)

\bibitem{shrivastava2011data}
Shrivastava, A., Malisiewicz, T., Gupta, A., Efros, A.A.: Data-driven visual
  similarity for cross-domain image matching. In: Proceedings of the 2011
  SIGGRAPH Asia Conference (2011)

\bibitem{singh2012unsupervised}
Singh, S., Gupta, A., Efros, A.A.: Unsupervised discovery of mid-level
  discriminative patches. In: Proceedings of the European Conference on
  Computer Vision (2012)

\bibitem{pwcnet}
Sun, D., Yang, X., Liu, M.Y., Kautz, J.: Pwc-net: Cnns for optical flow using
  pyramid, warping, and cost volume. In: Proceedings of the IEEE Conference on
  Computer Vision and Pattern Recognition (2018)

\bibitem{Szeliski2006}
Szeliski, R.: Image alignment and stitching: A tutorial. Found. Trends. Comput.
  Graph. Vis.  (2006)

\bibitem{yfcc100m}
Thomee, B., Shamma, D.A., Friedland, G., Elizalde, B., Ni, K., Poland, D.,
  Borth, D., Li, L.J.: Yfcc100m: The new data in multimedia research.
  Communications of the ACM  (2016)

\bibitem{tian2017l2}
Tian, Y., Fan, B., Wu, F.: L2-net: Deep learning of discriminative patch
  descriptor in euclidean space. In: Proceedings of the IEEE Conference on
  Computer Vision and Pattern Recognition (2017)

\bibitem{ulyanov2018deep}
Ulyanov, D., Vedaldi, A., Lempitsky, V.: Deep image prior. In: Proceedings of
  the IEEE Conference on Computer Vision and Pattern Recognition (2018)

\bibitem{videoCor}
Vondrick, C., Shrivastava, A., Fathi, A., Guadarrama, S., Murphy, K.: Tracking
  emerges by colorizing videos. In: Proceedings of the European Conference on
  Computer Vision (2018)

\bibitem{timeCycle}
Wang, X., Jabri, A., Efros, A.A.: Learning correspondence from the
  cycle-consistency of time. In: The IEEE Conference on Computer Vision and
  Pattern Recognition (2019)

\bibitem{oaflow}
Wang, Y., Yang, Y., Yang, Z., Zhao, L., Wang, P., Xu, W.: Occlusion aware
  unsupervised learning of optical flow. In: Proceedings of the IEEE Conference
  on Computer Vision and Pattern Recognition (2018)

\bibitem{wang2004}
Wang, Z., Bovik, A.C., Sheikh, H.R., Simoncelli, E.P.: Image quality
  assessment: from error visibility to structural similarity. IEEE Transactions
  on Image Processing  (2004)

\bibitem{wulff2017optical}
Wulff, J., Sevilla-Lara, L., Black, M.J.: Optical flow in mostly rigid scenes.
  In: Proceedings of the IEEE Conference on Computer Vision and Pattern
  Recognition (2017)

\bibitem{yin2018geonet}
Yin, Z., Shi, J.: Geonet: Unsupervised learning of dense depth, optical flow
  and camera pose. In: Proceedings of the IEEE Conference on Computer Vision
  and Pattern Recognition (2018)

\bibitem{oanet}
Zhang, J., Sun, D., Luo, Z., Yao, A., Zhou, L., Shen, T., Chen, Y., Quan, L.,
  Liao, H.: Learning two-view correspondences and geometry using order-aware
  network. Proceedings of the IEEE International Conference on Computer Vision
  (2019)

\bibitem{antialiased}
Zhang, R.: Making convolutional networks shift-invariant again. In:
  International Conference on Machine Learning (2019)

\bibitem{zhou2017scene}
Zhou, B., Zhao, H., Puig, X., Fidler, S., Barriuso, A., Torralba, A.: Scene
  parsing through ade20k dataset. In: Proceedings of the IEEE Conference on
  Computer Vision and Pattern Recognition (2017)

\bibitem{zhou2018semantic}
Zhou, B., Zhao, H., Puig, X., Xiao, T., Fidler, S., Barriuso, A., Torralba, A.:
  Semantic understanding of scenes through the ade20k dataset. International
  Journal on Computer Vision  (2018)

\bibitem{zhou2015flowweb}
Zhou, T., Jae~Lee, Y., Yu, S.X., Efros, A.A.: Flowweb: Joint image set
  alignment by weaving consistent, pixel-wise correspondences. In: Proceedings
  of the IEEE Conference on Computer Vision and Pattern Recognition (2015)

\bibitem{zhou2016learning}
Zhou, T., Krahenbuhl, P., Aubry, M., Huang, Q., Efros, A.A.: Learning dense
  correspondence via 3d-guided cycle consistency. In: Proceedings of the IEEE
  Conference on Computer Vision and Pattern Recognition (2016)

\end{thebibliography}
	
	\appendix
	\section*{Appendix}
	\addcontentsline{toc}{section}{Appendices}
	\renewcommand{\thesubsection}{\Alph{subsection}}
	
	\subsection{Dependency on $\matcheta$ and $\flowtheta$}
	\label{sec:app_param}
	Our training has 3 stages (Sec.~\ref{subsec:pipeline}): the model was firstly learned with the reconstruction loss $\losssim$ then added cycle-consistent flow loss $\losscycle$ and finally trained with all the losses (Equation~\ref{eq:totalloss}). In Table~\ref{tab:flowmatch}, we provide an analysis on the weighting parameters $\matcheta$ and $\flowtheta$ on sparse correspondences evaluation on MegaDepth~\cite{megaDepth} and report the accuracy at 3 pixels. We can see the stage 2 is not very sensitive with respect to $\flowtheta$ (Table~\ref{tab:flowtheta}), while the stage 3 with adding the mask loss is slightly more sensitive (Table~\ref{tab:matcheta}). Note that we then use the same parameters for fine-tuning on the different datasets. 
	
	\captionsetup[table]{farskip=2pt,captionskip=1pt,aboveskip=4pt}
	\begin{table}[h]
		\centering
		\caption{Dependency on $\matcheta$ and $\flowtheta$, we evaluate on sparse correspondences on MegaDepth~\cite{megaDepth} and report the accuracy at 3 pixels. (a) Training stage 2: dependency on $\flowtheta$ with $\matcheta$ = 0; (b) Training stage 3: dependency on $\matcheta$ with $\flowtheta$ = 1 (optimal in Table~\ref{tab:flowtheta}).}
		\begin{minipage}{0.5\textwidth}
			\centering
			\subfloat[(a) Training stage 2: dependency on $\flowtheta$ with $\matcheta$ = 0.]{
				\begin{tabular}{c||c}
					$\flowtheta$ & Acc. ($\le$ 3 pixels, MegaDepth~\cite{megaDepth})
					\\ \hline 
					2& 78.2 \\
					1& \bf 78.3 \\
					0.5& \bf 78.3 \\
				\end{tabular}~\label{tab:flowtheta}}
		\end{minipage}\hfill
		\begin{minipage}{0.5\textwidth}
			\centering
			\subfloat[(b) Training stage 3: dependency on $\matcheta$ with $\flowtheta$ = 1 (optimal in Table~\ref{tab:flowtheta}).]{
				\begin{tabular}{c||c}
					$\matcheta$ & Acc. ($\le$ 3 pixels, MegaDepth~\cite{megaDepth})
					\\ \hline 
					0.02& 83.0 \\
					0.01& \bf 83.5 \\
					0.005& 80.5 \\
				\end{tabular}\label{tab:matcheta}}
		\end{minipage}
		\label{tab:flowmatch}
	\end{table}
	
\end{document}